\documentclass[journal]{IEEEtran}
\ifCLASSINFOpdf
\else
\fi
\usepackage{amsmath}
\usepackage{array}

\usepackage{graphicx}
\usepackage{multirow}
\usepackage{booktabs}
\usepackage{array}
\usepackage{soul,color}
\usepackage{tablefootnote}
\usepackage{esvect}
\usepackage{comment}
\usepackage{booktabs}
\usepackage{adjustbox}
\usepackage{caption}
\usepackage{subcaption}
\usepackage{adjustbox}
\usepackage{xcolor}
\usepackage{cite}
\usepackage{longtable}
\usepackage{supertabular}
\usepackage{amssymb}
\usepackage{nicefrac}
\usepackage{amsmath,scalerel}

\begin{document}

\title{A Survey on Graph-Based Deep Learning \\ for Computational Histopathology}

\author{
David Ahmedt-Aristizabal,
Mohammad Ali Armin, 
Simon Denman,
Clinton Fookes,
Lars Petersson
\thanks{D. Ahmedt-Aristizabal, A. Armin and L. Petersson are with the Imaging and Computer Vision group, CSIRO Data61, Canberra, Australia. 
({Corresponding author: \tt\footnotesize david.ahmedtaristizabal@data61.csiro.au})}
\thanks{D. Ahmedt-Aristizabal, S. Denman and C. Fookes are with SAIVT, Queensland University of Technology, Brisbane, Australia. 
}
}

%



\maketitle

\begin{abstract}
With the remarkable success of representation learning for prediction problems, we have witnessed a rapid expansion of the use of machine learning and deep learning for the analysis of digital pathology and biopsy image patches. 
However, learning over patch-wise features using convolutional neural networks limits the ability of the model to capture global contextual information and comprehensively model tissue composition. The phenotypical and topological distribution of constituent histological entities play a critical role in tissue diagnosis.
As such, graph data representations and deep learning have attracted significant attention for encoding tissue representations, and capturing intra- and inter- entity level interactions.
In this review, we provide a conceptual grounding for graph analytics in digital pathology, including entity-graph construction and graph architectures, and present their current success for tumor localization and classification, tumor invasion and staging, image retrieval, and survival prediction. 
We provide an overview of these methods in a systematic manner organized by the graph representation of the input image, scale, and organ on which they operate.
We also outline the limitations of existing techniques, and suggest potential future research directions in this domain.
\end{abstract}


\IEEEpeerreviewmaketitle

\section{Introduction}

\IEEEPARstart{R}{ecent} advances in deep learning techniques have rapidly transformed these approaches into the methodology of choice for analyzing medical images, and in particular for histology image classification problems~\cite{deng2020deep}. Because of the increasing availability of large scale high-resolution whole-slide images (WSI) of tissue specimens, digital pathology and microscopy have become appealing application areas for deep learning algorithms.
Given wide variations in pathology and the often time-consuming diagnosis process, clinical experts have begun to benefit from computer-aided detection and diagnosis methods capable of learning  features that optimally represent the data~\cite{shen2017deep}.
This thorough survey serves as an accurate guide to biomedical engineering and clinical research communities interested in discovering the tissue composition-to-functionality relationship using image-to-graph translation and deep learning.

There are several review papers available that analyse the benefits of deep learning for providing reliable support for microscopic and digital pathology diagnosis and treatment decisions~\cite{van2021deep,srinidhi2020deep,deng2020deep,litjens2017survey,xing2017deep}, and specifically for cancer diagnosis~\cite{he2020deep}.
Compared to other medical fields such as dermatology, ophthalmology, neurology, cardiology, and radiology, digital pathology and microscopy is one of the most dominant medical applications of deep learning. One driving force behind innovation in computational pathology has been the introduction of grand challenges
(\textit{e.g.} NuCLS~\cite{amgad2021nucls}, BACH~\cite{aresta2019bach}, MoNuSeg~\cite{kumar2019multi}). Developed techniques that offer decision support to human pathologists have shown bright prospects for detecting, segmenting, and classifying the cell and nucleus; and detecting and classifying diseases such as cancer.

\begin{figure}[!t]
\centering
\includegraphics[width=0.95\linewidth]{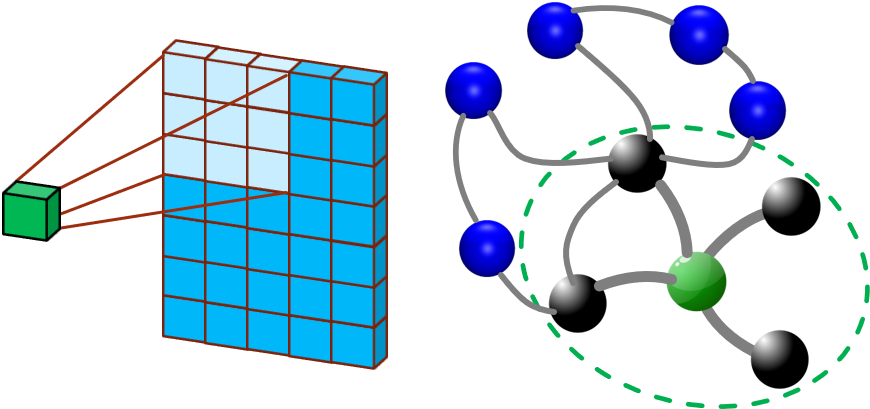}
\caption{
Traditional CNNs excel at modelling local relations in grid representation, where the topology of the neighborhood is constant (Left).
GCNs can take into account different neighbouring relations (global relation) by going beyond the local pixel neighbourhoods used by convolutions. On a graph, the neighbours of a node are unordered and variable in size (Right).}
\label{fig:Fig1a}
\vspace{-10pt}
\end{figure}

Deep learning techniques such as convolutional neural networks (CNNs) have demonstrated success in extracting image-level representations, however, they are inefficient when dealing with relation-aware representations. Modern deep learning variations of graph neural networks (GNNs) have made a significant impact in many technological domains for describing relationships.
Graphs, by definition, capture relationships between entities and can thus be used to encode relational information between variables~\cite{wu2020comprehensive}. As a result, special emphasis has been placed on the generalisation of GNNs into non-structured and structured scenarios.
Traditional CNNs analyse local areas based on fixed connectivity (determined by the convolutional kernel), leading to limited performance, and difficulty in interpreting the structures being modeled.
Graphs, on the other hand, offer more flexibility to analyse unordered data by preserving neighboring relations. This difference is illustrated in Fig.{~\ref{fig:Fig1a}}.

\begin{figure}[!t]
\centering
\includegraphics[width=1\linewidth]{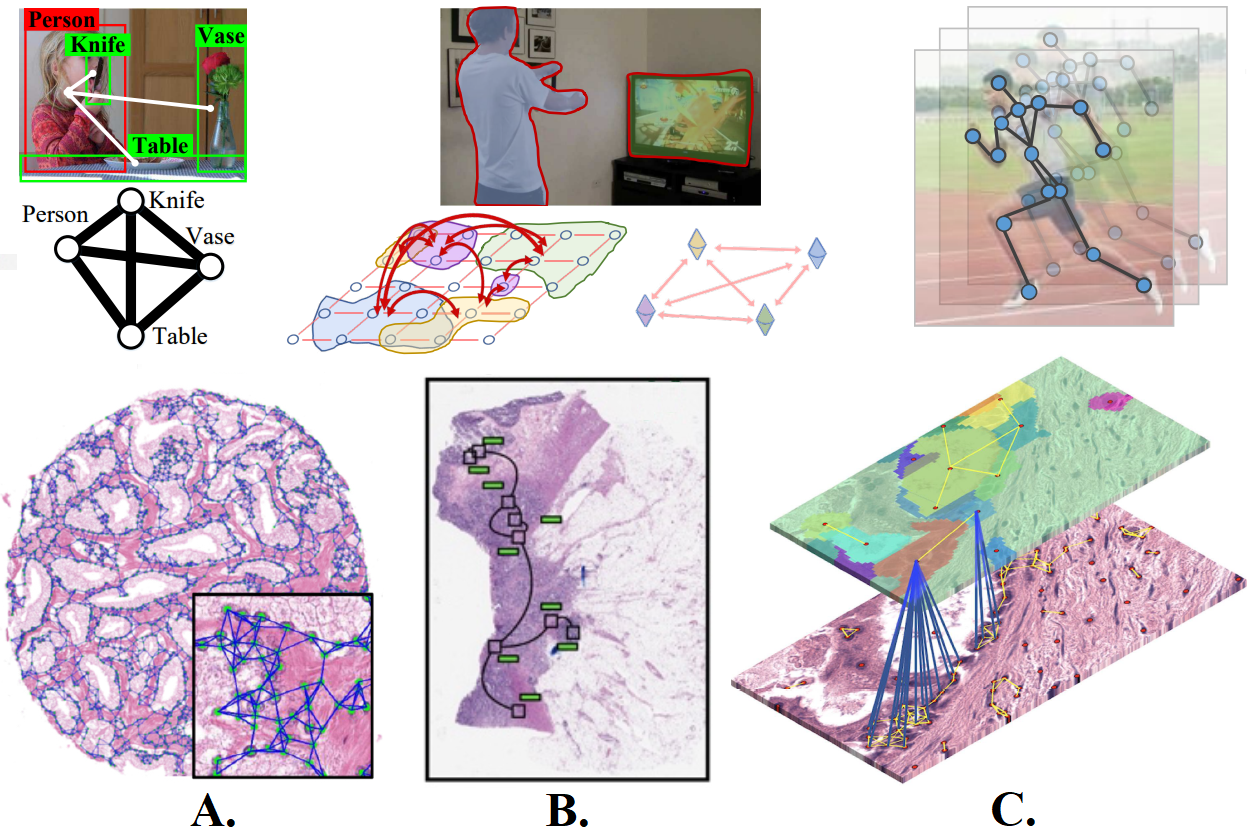}
\vspace{-15pt}
\caption{
\textbf{Top:} 
Graph-based representation of images for relation-aware human-object interaction, image segmentation, and human pose estimation (left-to-right). Images adapted from~\cite{qi2018learning,chen2019graph,li2020temporal}.
\textbf{Bottom:} 
\textbf{A.} Cell-graph representation for prostate cancer.
\textbf{B.} Tissue-graph representation for colorectal cancer.
\textbf{C.} Hierarchical cell-to-tissue graph representation for breast cancer.
Images adapted from~\cite{wang2020weakly,levy2020topological,pati2020hact}.}
\label{fig:Fig1}
\vspace{-10pt}
\end{figure}

The adaptation of deep learning from images to graphs has received increased attention, leading to a new cross-domain field of graph-based deep learning which seeks to learn informative representations of graphs in an end-to-end manner. This field has exhibited remarkable success for various tasks as discussed by recent surveys on graph deep learning frameworks and their applications~\cite{makarov2021survey,georgousis2021graph,wu2020comprehensive,zhang2020deep}. 
Graph embeddings have appeared in computer vision tasks where graphs can efficiently define relationships between objects, or for the purpose of graph-structured image analysis. 
Interesting results have been obtained for object detection, semantic segmentation, skeleton-based action recognition, image classification and human-object interaction tasks as illustrated in Fig.~\ref{fig:Fig1} (Top). 

Medical applications have benefited from rapid progress in the field of computer vision and GNNs. 
The development of GNNs has seen the application of deep learning methods to GNNs, such as graph convolutional networks (GCNs). These models have been proposed as a powerful tool to model functional and anatomical structures, brain electrical activity, and segmentation of the vasculature system and organs~\cite{ahmedt2021survey1}. 

Histological images depict the micro-anatomy of a tissue sample, and pathologists use histological images to make diagnoses based on morphological changes in tissues, the spatial relationship between cells, cell density, and other factors. Graph-based methods, which can capture geometrical and topological properties, are able to model cell-level information and overall tissue micro-architecture. 
Prior to the advent of deep learning, numerous approaches for processing histopathological images as graphs were investigated~\cite{sharma2015review}. 
These methods used classical machine learning approaches, which are less accurate for graph classification compared to GCNs. 
The capabilities of graph-based deep learning, which bridges the gap between deep learning methods and traditional cell graphs for disease diagnosis, are yet to be sufficiently investigated.

In this survey, we analyse how graph embeddings are employed in histopathology diagnosis and analysis. 
While graphs are not directly expressed within this data, they can efficiently describe relationships between tissue regions and cells. 
This setting offers a very different task for GNNs in comparison to analysis of unstructured data such as electrophysiological and neuroimaging recordings where the data can be directly mapped to a graph{~\cite{ahmedt2021survey1}}.
Selected samples of graph representations in digital pathology (cell-graph, patch-graph, tissue-graph and cell-tissue representation) used to capture and learn relevant morphological regions that will be covered in this review are illustrated in Fig.{~\ref{fig:Fig1}} (Bottom).

This survey offers a comprehensive overview of preprocessing, graph models and explainability tools used in computational pathology, highlighting the capability of GNNs to detect and associate key tissue architectures, regions of interest, and their interdependence.
Although some papers have surveyed conventional cell graphs with handcrafted features to characterize the entities~\cite{sharma2015review,irshad2013methods}, and others have briefly touched upon the benefits of GCNs in biology and medicine{~\cite{li2021representation}}, to the best of our knowledge, no systematic review exists that presents and discusses all relevant works concerning graph-based representations and deep learning models for computational pathology.

\vspace{-6pt}
\subsection{Why graph-based deep learning for characterizing diseases through histopathology slides?}

Deep learning has increased the potential of medical image analysis by enabling the discovery of morphological and textural representations in images solely from the data. Although CNNs have shown impressive performance in the field of histopathology analysis, they are unable to capture complex neighborhood information as they analyse local areas determined by the convolutional kernel. To extract interaction information between objects, a CNN needs to reach sufficient depth by stacking multiple convolutional layers, which is inefficient. This leads to limitations in the performance and interpretability of the analysis of anatomical structures and microscopic samples.

Graph convolutional networks (GCNs) are a deep learning-based method that operate over graphs, and are becoming increasingly useful for medical diagnosis and analysis~\cite{ahmedt2021survey1}. GCNs can better exploit irregular relationships and preserve neighboring relations compared with CNN-based models~\cite{wu2020comprehensive}. 
Below we outline the reasons why current research in histopathology has shifted the analytical paradigm from pixel to entity-graph processing:

\begin{enumerate}
    \item The potential correlations among images are ignored during traditional CNN feature learning, however, a GCN can be introduced to estimate the dependencies between images and enhance the discriminative ability of CNN features~\cite{Shi2019GraphCN}.

    \item CNNs have been commonly used for the analysis of whole slide images (WSI) by classifying fixed-sized biopsy image patches using fixed fusion rules such as averaging features or class scores, or weighted averaging with learnable weights to obtain an image-level classification score. 
    Aggregation using a CNN also includes excessive whitespace, putting undue reliance on the orientation and location of the tissue segment.
    Even though CNN-based models have practical merits through considering important patches for prediction, they dismiss the spatial relationships between patches, or global contextual information.
    Architectures are required to be capable of dealing with size and shape variation in region-of-interests (ROIs), and must encode the spatial context of individual patches and their collective contribution to the diagnosis, which can be addressed with graph-based representations~\cite{aygunecs2020graph,raju2020graph}.
    
    \item A robust computer-aided detection system should be able to capture multi-scale contextual features in tissues, which can be difficult with traditional CNN-based models. A pathological image can be transformed into a graph representation to capture the cellular morphology and topology (cell-graph)~\cite{anand2020histographs}, and the attributes of the tissue parts and their spatial relationships (tissue-graph)~\cite{zhang2020ms,pati2020hact}.
    
    \item Graph representations can enhance the interpretation of the final representation by modeling relations among different regions of interest. Graph-based models offer a new way to verify existing observations in pathology. 
    Attention mechanisms with GCNs, for example, highlight informative nuclei and inter-nuclear interactions, allowing the production of interpretable maps of tissue images displaying the contribution of each nucleus and its surroundings to the final diagnosis~\cite{sureka2020visualization}.
    
    \item By incorporating any task-specific prior pathological information, an entity-graph can be customized in various ways. As a result, pathology-specific interpretability and human-machine co-learning are enabled by the graph format~\cite{pati2021hierarchical}.
    
    \item GCNs are a complimentary method to CNNs for morphological feature extraction, and they can be employed instead of, or in addition to CNNs during multimodal fusion for fine-grained patient stratification~\cite{chen2020pathomic}.
    
\end{enumerate}

\vspace{-6pt}
\subsection{Contribution and organisation}

Compared to other recent reviews on traditional deep learning in histopathology slides, our manuscript captures the current efforts relating to entity-graphs and recent advancements in GCNs for characterizing diseases and pathology tasks.

Papers included in the survey are obtained from various journals, conference proceedings and open-access repositories.
Table~\ref{table:Applications} outlines the applications that were addressed across all reviewed publications. It is noted that breast cancer analysis constitutes the major application in digital pathology that has been analyzed using graph-based deep learning techniques.

This review is divided into three major sections.
In Section~\ref{sec:sec2} we provide a technical overview of the prevailing tools for entity-graph representation and graph architectures used in accelerating digital pathology research.
In Section~\ref{sec:sec3} we introduce the current applications of deep graph representation learning and cluster these proposals based on the graph construction (cell-graph, patch-graph, tissue-graph, hierarchical graph) and feature level fusion methods followed by the task or organ on which they operate.
Finally, Section~\ref{sec:sec4} highlights open problems and perspectives regarding the shifting analytical paradigm from pixel to entity-based processing. Specifically, we discuss the topics of graph construction, embedding expert knowledge, complexity of graph models, training paradigms, and graph model interpretability.

\begin{table}[t!]
\caption{Summary of applications of graph-based deep learning in histopathology covered in this survey.}
\vspace{-2pt}
\centering
\label{table:Applications}
\resizebox{0.44\textwidth}{!}{%
\begin{tabular}{
lcl
}
\toprule
\textbf{Application} & \textbf{\#Applications} & \textbf{Reference} \\
\midrule
Breast cancer & 11
& \cite{jaume2020quantifying,jaume2020towards,sureka2020visualization,anand2020histographs,ozen2021self,lu2020capturing,aygunecs2020graph,ye2019improving,pati2021hierarchical,pati2020hact,zhang2020ms} \\
Colorectal cancer & 6
& \cite{studer2021classification,zhou2019cgc,zhao2020predicting,raju2020graph,ding2020feature,levy2020topological} \\
Prostate cancer & 3  
& \cite{wang2020weakly,anklin2021learning,sureka2020visualization} \\
Lung cancer & 3  
& \cite{adnan2020representation,zheng2019encoding,li2018graph} \\
Cervical cancer & 2 
& \cite{shi2020cervical,Shi2019GraphCN} \\
Lymphoma & 1 
& \cite{levy2020topological} \\
Skin cancer & 1 
& \cite{wu2019weakly} \\
Renal cancer & 1  
& \cite{chen2020pathomic} \\
\midrule
\textbf{Total} & \textbf{28} 
& \\
\bottomrule
\end{tabular}}
\vspace{-9pt}
\end{table}

\begin{figure*}[!t]
\centering
\includegraphics[width=1\linewidth]{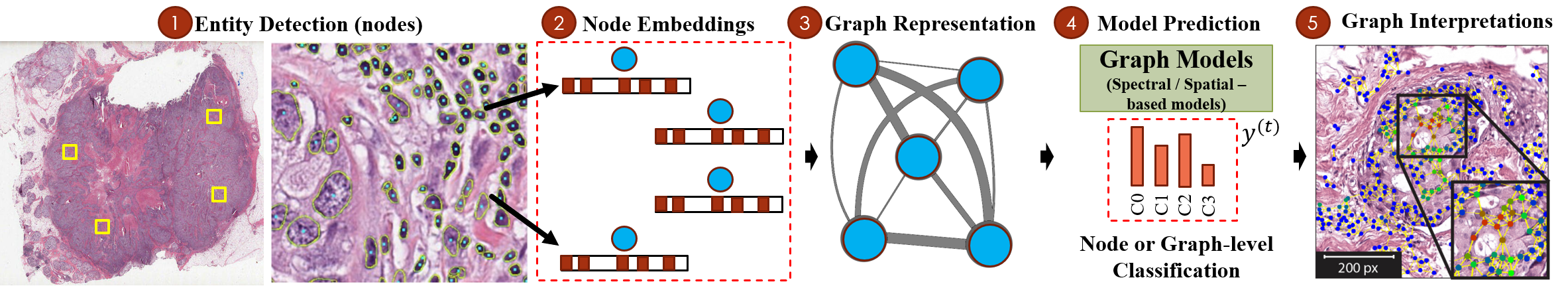}
\caption{
Overview of a standard graph-based workflow in computational pathology.
The WSI image is first transformed into one or more graphs.
\textbf{1.} The entities can be nuclei, patches or tissue regions. 
\textbf{2.} Node features comprise handcrafted or deep learning features to characterize the entities.
\textbf{3.} The edges encode intrinsic relationships (spatial or semantic) among the entities.
\textbf{4.} Graph encoding and classification (node-level or graph-level prediction): the graph representation is processed using GNNs and its variants such as ChebNet, GCN, GraphSAGE, GAT, and GIN, including different graph pooling strategies (global or hierarchical pooling).
\textbf{5.} Graph interpretations: a set of GNN model interpretability tools such as graph attentions or post-hoc graph explainers (\textit{e.g.}~GNNExplainer and GraphGrad-CAM.)}
\label{fig:Fig2}
\vspace{-8pt}
\end{figure*}

%
%

\section{Graph representation learning \\ in digital pathology: Background}
\label{sec:sec2}%

Translating patient histopathological images into graphs to encode the spatial context of cells and tissues for a given patient has been used to improve prediction accuracy of various pathology tasks. Graph representations followed by GNN-based models and interpretability approaches allows pathologists to directly comprehend and reason for the outcomes. GNNs can also serve a variety of prediction purposes by adapting different designs, such as performing node-level and graph-level predictions.

A standard entity-graph based pathological workflow requires several phases, such as node and graph topology definition, as well as the choice of GNN architecture.
In this section, we provide technical insights of these phases that are required for graph analytics in computational pathology: 
(1) Graph representation (entity, embeddings and edges definition);
(2) Graph models (graph structures for processing graph-structured);
and (3) Explainability (a set of interpretation methodologies such as model-based and post-hoc interpretability).
A traditional framework with aforementioned phases is illustrated in Fig.{~\ref{fig:Fig2}}.
A deep analysis of each GNN model can be found in survey papers that deal with graph architectures~\cite{wu2020comprehensive,zhang2020deep}.




\vspace{-6pt}
\subsection{Histopathology graph representation}

\subsubsection{Preliminaries}

A graph can be represented by $G=(\mathcal{V},\mathcal{E},W)$, where $V$ is a vertex set with $|\mathcal{V}|=n$ nodes and $\mathcal{E}$ denotes the set of edges connecting these nodes.
Data in $\mathcal{V}$ can be represented by a feature matrix $\mathrm{X} \in \mathbb{R}^{n \times d}$, where $n$ and $d$ denote the input feature dimensions.
$W \in \mathbb{R}^{n \times n}$ is a binary or weighted adjacency matrix describing the connections between any two nodes in $\mathcal{V}$, in which the importance of the connections between the \textit{i}-th and the \textit{j}-th nodes is measured by the entry  $W$ in the \textit{i}-th row and \textit{j}-th column, and denoted $w_{ij}$.
Commonly used methods to determine the entries, $w_{ij}$, of $W$ include Pearson correlation-based graph, the K-nearest neighbor (KNN) method, and the distance-based graph~\cite{shuman2013emerging}. 
In general, GNNs learn a feature transformation function for $\mathrm{X}$ and produce output $Z \in \mathbb{R}^{n \times d^{'}}$ , where $d^{'}$ denotes the output feature dimension.

Presented graph methods in digital pathology typically use data in one of two forms. Whole slide images (WSI), also known as virtual microscopy, are high-resolution images generated by combining many smaller image tiles or strips and tiling them to form a single image. Tissue microarrays (TMAs) consist of paraffin blocks produced by extracting cylindrical tissue cores and inserting them into a single recipient block (microarray) in a precisely spaced pattern. With this technique, up to 1000 tissue cores can be assembled in a single paraffin block to allow multiplex histological analysis.

\subsubsection{Graph construction}

Graph representations have been used in digital pathology for multiple tasks where a histology image is described as an entity-graph, and nodes and edges of a graph denote biological entities and inter-entity interactions respectively. 
The entities can be biologically-defined such as nuclei and tissue regions, or can be defined patch-wise. 
Therefore, constructing an entity-graph for graph analytics in computational pathology demands the following pre-processing steps.

\paragraph{Node definition}

WSI usually includes significant non-tissue regions. To identify tissue regions the foreground is segmented with Gaussian smoothing and OTSU thresholding{~\cite{otsu1979threshold}}.

One of the most common graph representation, cell-graphs, requires model training and fine-tuning for cell detection or segmentation. To detect nuclei several methods have been used such as Hover-Net{~\cite{graham2019hover}}, CIA-Net{~\cite{zhou2019cia}}, 
UNet{~\cite{ronneberger2015u}} and cGANs{~\cite{mahmood2019deep}}, that are trained on multi-organ nuclei segmentation datasets (MoNuSeg{~\cite{kumar2017dataset}}, PanNuke{~\cite{gamper2019pannuke}}, CoNSep{~\cite{graham2019hover}}). The entities can also be calculated using agglomerative clustering{~\cite{mullner2011modern}} of detected cells.

The nodes in a graph can also be represented by fixed-sized patches (patch-graphs) randomly sampled from the raw WSI or by using a patch selection method where non-tissue regions are removed{~\cite{kalra2020yottixel}}. 
Important patches can be sampled from segmented tissues using color thresholds where patches with similar features (tissue cluster) are modeled as a node.
Pre-trained deep learning models on tissue datasets (\textit{e.g.} NCT-CRC-HE-100{~\cite{kather_jakob_nikolas}}) have also been used to detect the tumor region of the specific pathological task.

Meaningful tissue regions have been also used as nodes to capture the tissue distribution (tissue-graphs). To separate tissue structures, superpixels{~\cite{bejnordi2015multi}} obtained using unsupervised algorithms such as simple linear iterative clustering (SLIC){~\cite{achanta2012slic}}) become nodes.

\paragraph{Node embeddings}

Node features can comprise hand-crafted features including morphological and topological properties (\textit{e.g.} shape, size, orientation, nuclei intensity, and the chromaticity using the gray-level co-occurrence matrix).
For cell-graph representations, some works include learned features extracted from the trained model used to localise the nuclei.

In patch-graph methods, deep neural networks are used to automatically learn a feature representation from patches around the centroids of the nuclei and tissue regions. If the entity is larger than the specified patch size, multiple patches inside the entity are processed, and the final feature is computed as the mean of the patch-level deep features.
Some works have aggregated features from neighboring patches and combined them to obtain a central node representation to increase feature learning performance.
Authors have adopted CNNs (MobileNetV2, DenseNet, ResNet-18 or ResNet-50{~\cite{he2016deep}}), and encoder-decoder segmentation models (UNet{~\cite{ronneberger2015u}}) for the purpose of deep feature extraction. To generate patch-level embeddings, ImageNet-pretrained CNN as well as a CNN pretrained for tissue sub-compartment classification task have been used. 

\paragraph{Edge definition}

The edge configuration encodes the cellular or tissue interactions, \textit{i.e.} how likely two nearby entities will interact and consequently form an edge. 
This topology is often defined heuristically using a pre-defined proximity threshold, a nearest neighbor rule, a probabilistic model, or a Waxman model{~\cite{sharma2015review}}.
The graph topology can also be computed by constructing a region adjacency graph (RAG) {~\cite{potjer1996region}} by using the spatial centroids of superpixels.

\subsubsection{Training paradigms}

From the perspective of supervision, we can categorize graph learning tasks into different training settings. Such approaches have also been used to extract effective representations from data.

\begin{itemize}
    \item The \textit{Supervised} learning setting provides labeled data for training.

    \item \textit{Weakly or partially supervised learning} refers to models that are trained using examples that are only partially annotated.

    \item \textit{Semi-supervised learning} trains a model using a small set of annotated samples, then generates pseudo-labels for a large set of samples without annotations, and learns a final model by mixing both sets of samples.
    
    \item \textit{Self-supervised learning} is a form of unsupervised learning in which the data provides supervisory signals when learning a representation via a proxy task. Annotated data is used to fine-tune the representation once it has been learned.
    Some self-supervised approaches adopted as feature extractors include contrastive predictive coding (CPC){~\cite{oord2018representation}}, texture auto encoder (Deep Ten){~\cite{zhang2017deep}}, and variational autoencoders (VAE){~\cite{kingma2013auto}}.

\end{itemize}

\vspace{-6pt}
\subsection{Graph neural networks models}

Following graph building, the entity graph is processed using a graph-based deep learning model that works with graph-structured data to perform analysis.

GCNs can be broadly categorised as spectral-based~\cite{defferrard2016convolutional,kipf2017semi} and spatial-based~\cite{hamilton2017inductive}.
Spectral-based GCNs use spectral convolutional neural networks, that build upon the graph Fourier transform and the normalized Laplacian matrix of the graph. Spatial-based GCNs define a graph convolution operation based on spatial relationships that exist among graph nodes. 

Graph convolutional networks, similar to CNNs, learn abstract feature representations for each feature at a node via message passing, in which nodes successively aggregate feature vectors from their neighborhood to compute a new feature vector at the next hidden layer in the network.

A basic GNN consists of two components:
The \textit{AGGREGATE} operation can aggregate neighboring node representations of the center node, whereas the \textit{COMBINE} operation combines the neighborhood node representation with the center node representation to generate the updated center node representation.
The Aggregate and Combine at each $l-th$ layer of the GNN can be defined as follows:
\vspace{-2pt}
\begin{equation}
h_{\mathcal{N}_v}^{(t)} = \text{AGGREGATE}^{(l)} \left( \big\{ h_u^{l-1}, \forall u \in \mathcal{N}_v \big\} \right) ,
\end{equation}
where $h_{\mathcal{N}_v}^{(t)}$ is the aggregated node feature of the neighbourhood, $h_u^{l-1}$ is the node feature in neighbourhood $\mathcal{N}(\cdot)$ of node $v$.
\vspace{-2pt}
\begin{equation}
\resizebox{0.43\textwidth}{!}{$h_v^{(t)} = \text{COMBINE}^{(l)} \left( h_v^{t-1}, h_{\mathcal{N}_v}^{(t)} \right) =  \sigma (W^t \cdot [ h_v^{t-1} \| h_{\mathcal{N}_v}^t   ]   ) $}, 
\end{equation}
where $h_v^{(t)}$ is the node representation at the $l-th$ iteration. $h_v^{(0)} = x_{v}$ where $x_{v}$ is the initial feature vector for the node, $\sigma$ denotes the logistic sigmoid function, and $\|$ denotes vector concatenation.

\begin{figure}[!t]
\centering
\includegraphics[width=1\linewidth]{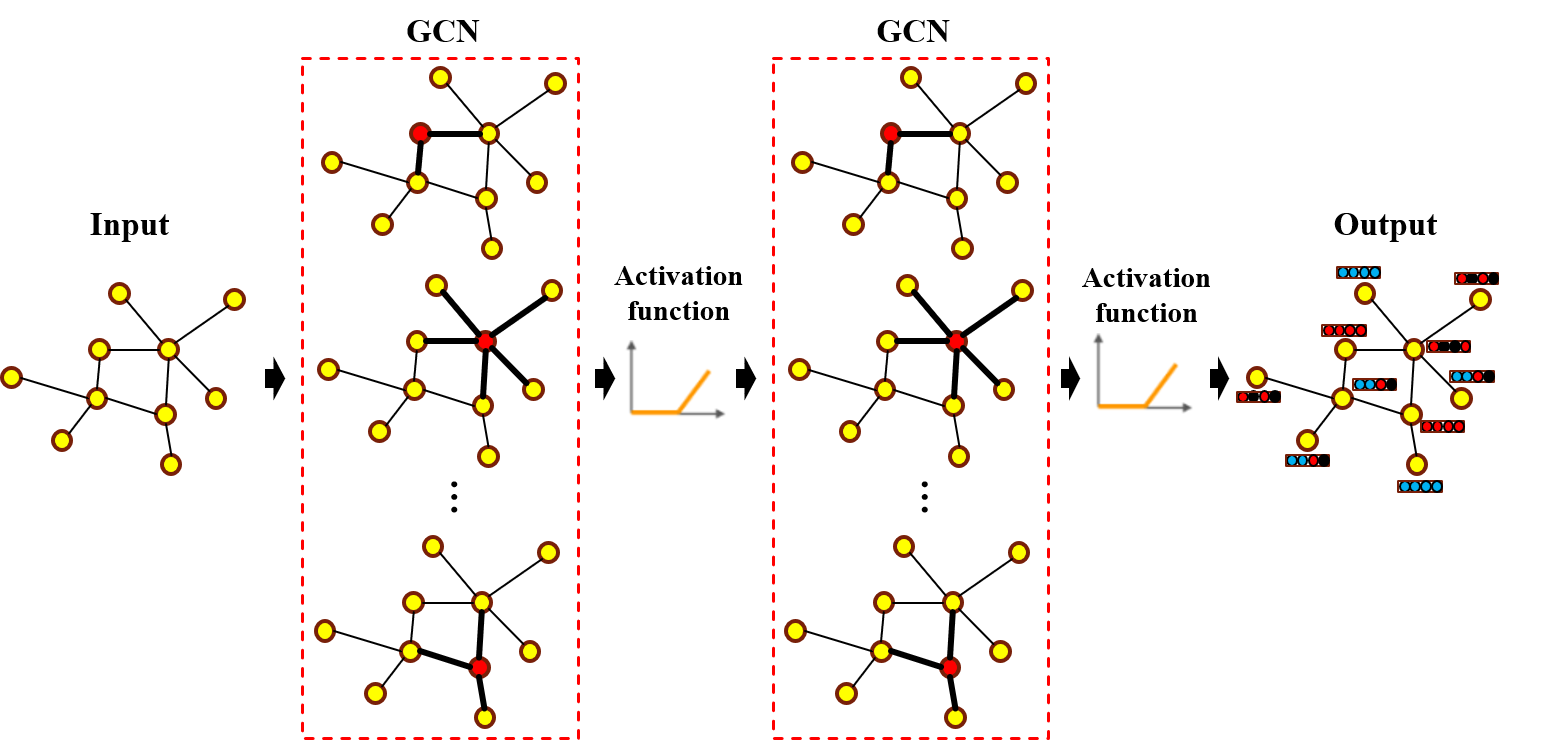}
\caption{
Representation of graph architectures for node-level classification. Recreated from{~\cite{wu2020comprehensive}}.
}
\label{fig:Fig2-1}
\vspace{-8pt}
\end{figure}

With the network structure and node content information as inputs, the outputs of GNNs can focus on various graph analytic tasks using one of the processes listed below:

\begin{itemize}
    \item \textit{Node-level prediction}: A GNN operating at the node-level computes values for each node in the graph and is thus useful for node classification and regression purposes.
    In node classification, the task is to predict the node label for every node in a graph. To compute the node-level predictions, the node embedding is input to a Multi-Layer Perceptron (MLP) (See Fig.{~\ref{fig:Fig2-1}}).

    \item \textit{Graph-level prediction}: Refers to GNNs that predict a single value for an entire graph. This is mostly used to classify entire graphs, or compute similarities between graphs.
    To compute graph-level predictions, the same node embedding used in node-level prediction is input to a pooling process followed by a separate MLP (See Fig.{~\ref{fig:Fig2-2}}).  

\end{itemize}

In the following subsections, we describe in more detail the GNN architectures considered in digital pathology analysis methods. Different GNN variants employ different aggregators to acquire information from each node's neighbors, as well as different techniques to update the nodes' hidden states. In GNNs, the number of parameters is dependent on the number of node and edge features, as their aggregation is learned.

\subsubsection{ChebNet}
The convolution operation for spectral-based GCNs is defined in the Fourier domain by determining the eigen decomposition of the graph Laplacian~\cite{bruna2013spectral}. 
The normalized graph Laplacian is defined as $L=I_N-D^{-1/2}AD^{-1/2}=U \Lambda U^T$ ($D$ is the degree matrix and $A$ is the adjacency matrix of the graph), where the columns of $U$ are the matrix of eigenvectors and $\Lambda$ is a diagonal matrix of its eigenvalues. The operation can be defined as the multiplication of a signal $x \in \mathbb{R}^N$ (a scalar for each node) with a filter $g_{\theta}=\text{diag}(\theta)$, parameterized by $\theta \in \mathbb{R}^N$,
\begin{equation}
g_{\theta} \star x =  U g_\theta (\Lambda) U^Tx.
\end{equation}

Defferrard et al.~\cite{defferrard2016convolutional} proposed a Chebyshev spectral CNN (ChebNet), which approximates the spectral filters by truncated Chebyshev polynomials, avoiding the calculation of the eigenvectors of the Laplacian matrix, and thus reducing the computational cost.
A Chebyshev polynomial $T_m(x)$ of order $m$ evaluated at $\tilde{L}$ is used. Thus the operation is defined as,
\vspace{-2pt}
\begin{equation}
g_{\theta} \star x \approx \sum_{m=0}^{M-1} \theta_m T_m (\tilde{L})x ,
\label{eq:eq2}
\end{equation}
where $\tilde{L}$ is a diagonal matrix of scaled eigenvalues defined as $\tilde{L}=\nicefrac{2L}{\lambda_{\text{max}}}-I_N$. $\lambda_{\text{max}}$ denotes the largest eigenvalue of $L$. The Chebyshev polynomials are defined as $T_m(x)=2xT_{k-1}(x)-T_{k-2}(x)$ with $T_0(x)=1$ and $T_1(x)=x$.

\subsubsection{GCN}
A GCN is a spectral-based GNN with mean pooling aggregation. Kipf and Welling~\cite{kipf2017semi} presented the GCN using a localized first-order approximation of ChebNet.
It limits the layer-wise convolution filter to $K=1$ and uses a further approximation of $\lambda \approx 2$, to avoid overfitting and limit the number of parameters. Thus, Equation{~\ref{eq:eq2}} can be simplified to,
\begin{equation}
g_{\theta} \star x \approx \theta_0^{'}x + \theta_1^{'}x (L-I_N)x = \theta_0^{'}x + \theta_1^{'} D^{-1/2}AD^{-1/2}x .
\end{equation}
Here, $\theta_0^{'}, \theta_1^{'}$ are two unconstrained variables. 
A GCN further assumes that $ \theta = \theta_0^{'} = -\theta_1^{'} $, leading to the
following definition of a graph convolution:
\vspace{-2pt}
\begin{equation}
g_{\theta} \star x \approx \theta (I_N + D^{-1/2} A D^{-1/2} ) x
\end{equation}

The definition to a signal $X \in \mathbb{R}^{N \times C}$ with $C$ input channels and $F$ filters for feature maps is generalized as follows,
\vspace{-2pt}
\begin{equation}
Z = \tilde{D}^{-1/2} \tilde{A}\tilde{D}^{-1/2} X \Theta ,
\label{eq:eq3}
\end{equation}
where $\Theta \in \mathbb{R}^{C \times F}$ is the matrix formed by the filter bank parameters, and $Z \in \mathbb{R}^{N \times F}$ is the signal matrix obtained by convolution.
From a spatial-based perspective, Equation{~\ref{eq:eq3}} is reformulated in{~\cite{gilmer2017neural}} as a message passing layer which updates the node's representation $x_i^{k}$ as follows:
\vspace{-2pt}
\begin{equation}
\begin{split}
m_i^{k+1} = \sum_{j \in N(i) \cup i }  \frac{x_j^k}{\sqrt{ |N(J)|  |N(i)|  }} , \\
x_i^{k+1} = \sigma (W^k m_i^{k+1}), 
\label{eq:eq4}
\end{split}
\end{equation}
where $m_i^k$ is the output of a message passing iteration, $|N(J)|$ and $ |N(i)|$ denote the node degree of node $j$ and $i$ respectively, $W^k$ denotes a layer-specific trainable weight matrix and $\sigma$ is a non-linearity function.

\subsubsection{GraphSAGE}
GraphSAGE is a spatial-GCN which uses a node embedding with max-pooling aggregation. Hamilton et al.~\cite{hamilton2017inductive} offer an extension of GCNs for inductive unsupervised representation learning with trainable aggregation functions instead of simple convolutions applied to neighborhoods as in a GCN. The authors propose a batch-training algorithm for GCNs to save memory at the cost of sacrificing time efficiency.
In~\cite{hamilton2017inductive} three aggregating functions are proposed: the element-wise mean, an LSTM, and max-pooling. 
The mean aggregator is an approximation of the convolutional operation from the transductive GCN framework~\cite{kipf2017semi}. An LSTM is adapted to operate on an unordered set by permuting the neighbors of the node. In the pooling aggregator, each neighbor's hidden state is fed through a fully-connected layer, and then a max-pooling operation is applied to the set of the node’s neighbors.
These aggregator functions are denoted as,
\begin{equation}
h_{\mathcal{N}_v}^t = \text{max} \big\{ \sigma ( W_{\text{pool}} h_u^{t-1} + b_{\text{pool}}), \forall u \in \mathcal{N}_v    \big\}  , 
\end{equation}
where $\mathcal{N}_v$ is the neighborhood set of node $v$, $W_{\text{pool}}$ and $b_{\text{pool}}$ are the parameters to be learned, and $\text{max}\{ \cdot \}$ is the element-wise maximum.
Hence, following the message passing formulation in Equation{~\ref{eq:eq4}}, the node representation is updated according to, 
\vspace{-2pt}
\begin{equation}
\begin{split}
m_i^{k+1} = MEAN_{j \in N(i) \cup i } ( x_j^k )  , \\
x_i^{k+1} = \sigma (W^k m_i^{k+1}), 
\end{split}
\end{equation}

\subsubsection{GAT}
Inspired by the self-attention mechanism{~\cite{vaswani2017attention}}, graph attention networks (GAT){~\cite{velivckovic2017graph}} incorporate the attention mechanism into the propagation steps by modifying the convolution operation. 
GAT is a spatial-GCN model that incorporates masked self-attention layers into graph convolutions and uses a neural network architecture to learn neighbor-specific weights.
Veli{\v{c}}kovi{\'c} et al.~\cite{velivckovic2017graph} constructed a graph attention network by stacking a single graph attention layer, $a$, which is a single-layer feed-forward neural network, parametrized by a weight vector $\vec{a} \in \mathbb{R}^{2F^{i}}$. The layer computes the coefficients in the attention mechanisms of the node pair $(i,j)$ by,
\begin{equation}
\alpha_{i,j} = \frac{ \text{exp} (\text{LeakyReLu} ( \vec{a}^T [W\vec{h}_i \mathbin\Vert W\vec{h}_j] ) ) }
{ \sum_{k \in N_i \mathbb{N} }  \text{exp} (\text{LeakyReLu} ( \vec{a}^T [W\vec{h}_i \mathbin\Vert  W\vec{h}_k] ) ) } ,
\end{equation}
where $\mathbin\Vert$ represents the concatenation operation. The attention layer takes as input a set of node features $h=\{\vec{h_1},\vec{h_2},...,\vec{h_N}\}, \vec{h_i} \in R^F$, where $N$ is the number of nodes of the input graph and $F$ the number of features for each node, and produces a new set of node features $h^{'}=\{\vec{h_1}^{'},\vec{h_2}^{'},...,\vec{h_N}^{'}\}, \vec{h_i}^{'} \in R^F$ as its output.
To generate higher-level features, as an initial step a shared linear transformation, parametrized by a weight matrix $W \in R^{F'*F}$, is applied to every node and subsequently a masked attention mechanism is applied to every node, resulting in the following scores,
\begin{equation}
e_{ij} = a ( W \vec{h_i}, W \vec{h_j} ),
\end{equation}
that indicates the importance of node $j^{'}s$ features to node $i$. The final output feature of each node can be obtained by applying a non-linearity, $\sigma$,
\begin{equation}
h_i^{'} = \sigma ( \sum_{j \in N_i} \alpha_{ij} Wh_j ).
\end{equation}

The layer also uses multi-head attention to stabilise the learning process. $K$ different attention heads are applied to compute mutually independent features in parallel, and then their features are concatenated.

The attention coefficients are used to update the node representation according to the following message passing formulation,
\begin{equation}
\begin{split}
m_i^{k+1} = \sum_{j \in N(i)} \alpha_{i,j}^k W^k x_j^k , \\
x_i^{k+1} = \sigma (\alpha_{i,j}^k W^k x_j^k + m_i^{k+1}),
\end{split}
\end{equation}

\subsubsection{GIN}
The graph isomorphism network (GIN)~\cite{xu2018powerful} is a spatial-GCN that aggregates neighborhood information by summing the representations of neighboring nodes. Isomorphism graph-based models are designed to interpret graphs with different nodes and edges.
The representation of node $i$ itself is then updated using a MLP,
\begin{equation}
\begin{split}
m_i^{k+1} = \sum_{j \in N(i)} x_j^k , \\
x_i^{k+1} = F((1+\epsilon)  \cdot x_i^k+m_i^{k+1}), 
\end{split}
\end{equation}
where $F$ is the MLP and $\epsilon$ is either a learnable parameter or fixed. GIN’s aggregation and readout functions are injective, and thus are designed to achieve maximum discriminative power~\cite{xu2018powerful}.

\subsubsection{Other GNN architectures in histopathology}
Other GNN architectures considered for entity-graph evaluation in digital pathology that were proposed by the surveyed works include:

\begin{itemize}
    \item \textit{Edge graph neural network (EGNN)}~\cite{studer2021classification,gong2019exploiting}:
    Edge features are included when leveraging the graph structure in the network. 

    \item \textit{Robust spatial filtering (RSF)}~\cite{sureka2020visualization,anand2020histographs,such2017robust}:  
    These spatial-based models are more flexible when dealing with heterogenous graphs as the graph inputs can be easily incorporated into the aggregation function.
    
    \item \textit{Adaptive GraphSAGE}~\cite{raju2020graph,zhou2019cgc}:
    Graph networks with the ability to more effectively learn the embedding feature between nodes, by using a learnable pattern to adaptively aggregate multi-level embedding features for each node.
    3
    \item \textit{Jumping Knowledge Network (JK-Net)}
    Xu et al.~\cite{xu2018representation} proposed the Jumping Knowledge (JK) approach to adaptively leverage, for each node, different neighborhood ranges to better represent feature.

    \item \textit{Feature-enhanced spatial-GCN (FENet)}~\cite{ding2020feature,xu2018powerful}:
    This model is proposed to analyse non-isomorphic graphs, distinct from isomorphic graphs which strictly share the same adjacency neighborhood matrix. The feature-enhance mechanism adaptively selects the node representation from different graph convolution layers. The model adopts sum-pooling to capture the full structural information of the entire graph representation.

    \item \textit{Multi-scale graph wavelet neural network (MS-GWNN)} ~\cite{zhang2020ms,xu2019graph}:
    This spectral model leverages the localization property of graph wavelets to perform multi-scale analysis with a variety of scaling parameters in parallel, offering high efficiency and good interpretability for graph convolution.
    
\end{itemize}

\begin{figure}[!t]
\centering
\includegraphics[width=1\linewidth]{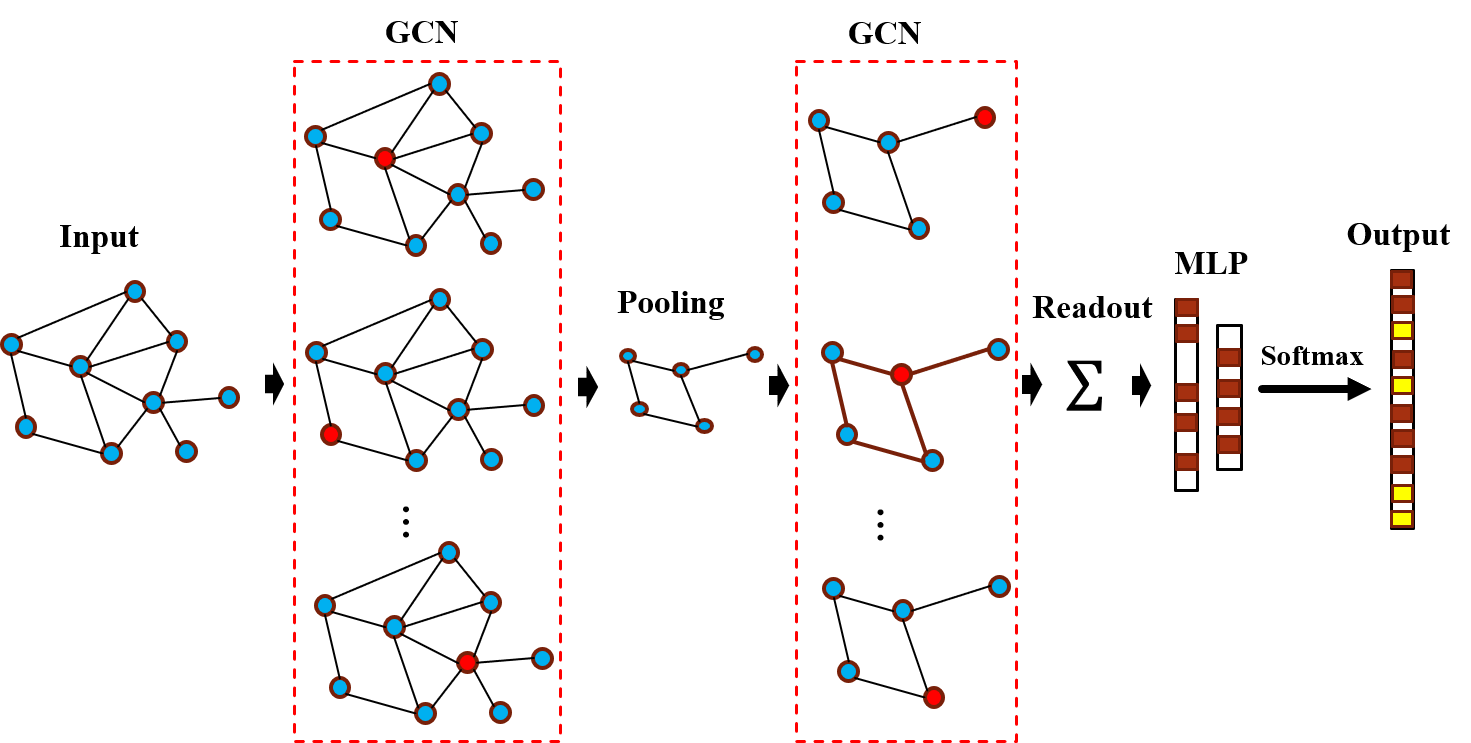}
\caption{
Representation of graph models for graph-level classification. Recreated from{~\cite{wu2020comprehensive}}.
}
\label{fig:Fig2-2}
\vspace{-8pt}
\end{figure}

\vspace{-6pt}
\subsection{Graph pooling}

Different graph pooling strategies have been developed to minimise the graph size in order to learn hierarchical features for improved graph-level classification, and reduce computational complexity.

\paragraph{Global pooling}
The most fundamental type of signal pooling on a graph is global pooling. It is also referred to as a readout layer in the literature. Similar to CNNs, mean, max, and sum functions are often utilized as basic pooling methods.
Other approaches, instead of employing these simple aggregators, transform the vertex representation to a permutation invariant graph-level representation or embedding. In particular, Li et al.{~\cite{li2015gated}} proposed a \textit{global attention pooling} system that uses a soft attention mechanism to determine which nodes are relevant to the present graph-level task and returns the pooled feature vector from all nodes.

\paragraph{Hierarchical pooling}
A graph pooling layer in the GCN pools information from multiple vertices to one vertex, to reduce graph size and expand the receptive field of the graph filters. Many graph classification methods use hierarchical pooling in conjunction with a final global pooling or readout layer to represent the graph as illustrated in Fig.{~\ref{fig:Fig2-2}}
Below we outline the most common hierarchical pooling techniques used in digital pathology. 

\begin{itemize}
    \item \textit{DiffPool:} Ying et al.{~\cite{ying2018hierarchical}} introduced the differentiable graph pooling operator (DiffPool) which uses another graph convolution layer to generate the assignment matrix for each node (\textit{i.e.} DiffPool does not simply cluster the nodes in a graph, but learns a cluster assignment matrix).
    
    \item \textit{SAGPool} The self-attention graph pooling (SAGPool) introduced by Lee et al.{~\cite{lee2019self}} is a hierarchical pooling method that performs local pooling operations over node embeddings in a graph. The pooling module considers both node features and graph topology and learns to pool features via a self-attention mechanism, which can reduce computational complexity.
\end{itemize}

\vspace{-6pt}
\subsection{Graph interpretations}

Graph representations embed biological entities and their interactions, but their explainability for digital pathology is less explored. While cells and their spatial interactions are visible in great detail, identifying relevant visual features is difficult. To undertake due diligence on model outputs and improve understanding of disease mechanisms and therapies, the medical community requires interpretable models.

The two most popular types of interpretation methodologies are model-based and post-hoc interpretability.
The former constrains the model so that it can quickly deliver meaningful details about the relationships that have been discovered (such as sparsity, modularity, etc). Here, internal model information such as weights or structural information can be accessed and used to infer group-level patterns across training instances.
The latter seeks to extract information about the learnt relationships in the model. These post-hoc methods are typically used to analyze individual feature input and output pairs, limiting their explainability to the individual sample level.

\begin{figure*}[!t]
\centering
\includegraphics[width=0.9\linewidth]{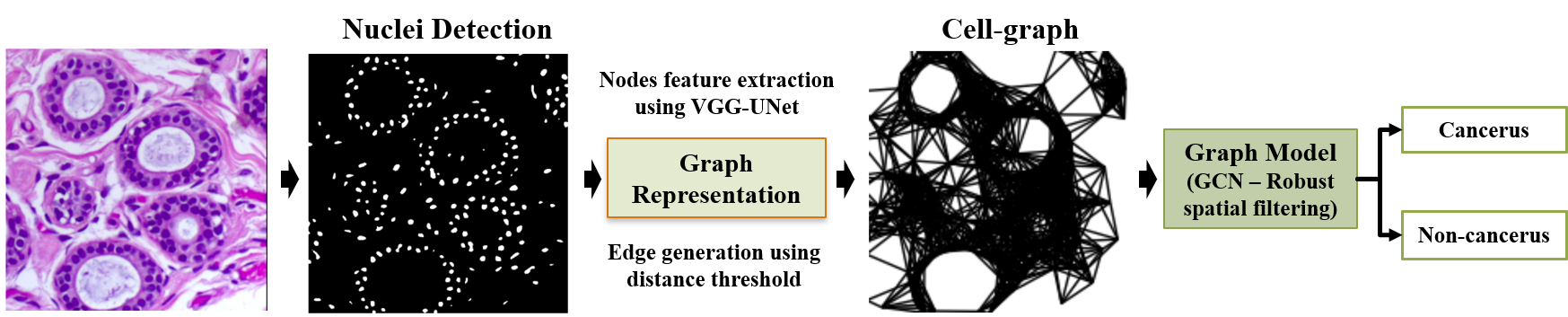}
\caption{
Cell-graph based representation. Nucleus detection is conducted using fully convolutional networks. Then, edge and vertex features are computed to obtain an entity-graph representation as input to a GCN for cancer classification. Recreated from~\cite{anand2020histographs}.
}
\label{fig:Fig3}
\vspace{-8pt}
\end{figure*}

\subsubsection{Attention mechanisms}

Graph-structured data can be both massive and noisy, and not all portions of the graph are equally important. As such, attention mechanisms can direct a network to focus on the most relevant parts of the input, suppressing uninformative features, reducing computational cost and enhancing accuracy. 
A gate-based attention mechanism{~\cite{arevalo2017gated}} controls, for example, the expressiveness of each feature.
Attention has also been used as an explanation technique where the attention weights highlight the nodes and edges in their relative order of importance, and can be used for discovering the underlying dependencies that have been learnt.
The activation map and gradient sensitivity of GAT models are used to interpret the salient input features at both the group and individual levels.

In a graph model with attention, selected layers of the graph are connected to an attention layer, and all attention layers are jointly trained with the network. A traditional attention mechanism that can be learned by gradient-based methods{~\cite{yang2016hierarchical}} can be formulated as,
\vspace{-4pt}
\begin{equation}
\begin{split}
u_t = \tanh ( W h_t + b), \\
\alpha_t = \dfrac{\exp(u_t^Tu_w)}{\sum_{j=1}^{n} \exp(u_t^Tu_w)} , \\
s_t = \sum_{t} \alpha_t h_t ,
\end{split}
\end{equation}
where $h_t$ is the output of a layer; and $W$, $u_w$ and $b$ are trainable weights and bias. The importance of each element in $h_t$ is measured by estimating the similarity between $u_t$ and $h_t$, which is randomly initialized. $\alpha_t$ is a softmax function. The scores are multiplied by the hidden states to calculate the weighted combination, $s_t$ (the attention-weighted final output).

\subsubsection{Graph explainers}

Several post-hoc feature attribution graph explainers have been presented in the literature including excitation backpropagation{~\cite{pope2019explainability}}, a node pruning-based explainer (GNNExplainer){~\cite{ying2019gnnexplainer}}, gradient-based explainers (GraphGrad-CAM{~\cite{selvaraju2017grad}} and GraphGrad-CAM++{~\cite{chattopadhay2018grad}}), a layerwise relevance propagation explainer (GraphLRP)~\cite{schwarzenberg2019layerwise,baldassarre2019explainability}, and deep graph mapper~\cite{bodnar2020deep}.

\begin{table*}[t]
\caption{Summary of applications and graphs models in computational pathology.}
\vspace{-5pt}
\centering
\label{table:pathology}
\resizebox{1\textwidth}{!}{%
\begin{tabular}{
l
>{\raggedright\arraybackslash}p{1.7cm}
>{\raggedright\arraybackslash}p{2.2cm}
c
>{\raggedright\arraybackslash}p{3cm} 
>{\raggedright\arraybackslash}p{10cm}}
\toprule
\textbf{Authors} &
\textbf{Topic} &
\textbf{Application} & 
\textbf{Entity-graph} & 
\textbf{GNN Model + Explainer} &  
\textbf{Input; Training (Node detection/embeddings); Training (GNN model/pathology task); Datasets; Additional remarks} \\
\midrule
%
Jaume et al. (2021)~\cite{jaume2020quantifying} & Classification & Breast cancer & CG &
GIN + Post-hoc explainers & WSI; Supervised; Supervised; BRACS~\cite{pati2020hact} (5 classes); Post-hoc explainers: GNNExplainer, GraphGrad-CAM, GraphGrad-CAM++, GraphLRP. \\ 
Jaume et al. (2020)~\cite{jaume2020towards} & Classification & Breast cancer & CG &
GIN + CGExplainer & WSI; Supervised; Supervised; BRACS~\cite{pati2020hact} (5 classes); Customized cell-graph explainer based on GNNExplainer. \\ 
Sureka et al. (2020)~\cite{sureka2020visualization} & Classification & Breast cancer / Prostate cancer & CG &
GCN, RSF + Attention/Node occlusion & WSI, TMAs; Supervised; Supervised; Breast cancer: BACH~\cite{aresta2019bach} (2 classes), Prostate cancer: TM~\cite{arvaniti2018automated} (2 classes); Gleason grade. \\
Anand et al. (2020)~\cite{anand2020histographs} & Classification & Breast cancer & CG &
GCN, RSF  & WSI; Supervised; Supervised; BACH~\cite{aresta2019bach} (4 classes). \newline \\
Studer et al. (2021)~\cite{studer2021classification} & Classification & Colorectal cancer & CG &
GCN, GraphSAGE, GAT, GIN, ENN, JK-Net & WSI; Supervised; Supervised; pT1-Gland Graph~\cite{studer2019graph} (2 classes); Graph-level output. Concatenation of global add, mean and max pooling). Dysplasia of intestinal glands. \\ 
Zhou et al. (2019)~\cite{zhou2019cgc} & Classification & Colorectal cancer & CG &
Adaptive GraphSAGE, JK-Net, Graph clustering & WSI; Supervised; Supervised; CRC dataset~\cite{awan2017glandular} (3 classes); Graph-level output. Hierarchical representation of cells based on graph clustering method from DiffPool). \\ 
Wang et al. (2020)~\cite{wang2020weakly} & Classification &  Prostate cancer & CG &
GraphSAGE, SAGPool & TMA; Self-supervised; Weakly-supervised; UZH prostate TMAs~\cite{zhong2017curated} (2 classes); Graph-level output. Grade classification (low and high-risk). \\
\midrule
%
Ozen et al. (2020)~\cite{ozen2021self} & ROI Retrieval &  Breast cancer & PG &
GCN, DiffPool & WSI; Supervised; Self-Supervised; Department of Pathology at Hacettepe University (private) (4 classes); Histopathological image retrieval (slide-level and ROI-level). \\
Lu et al. (2020)~\cite{lu2020capturing} & Classification & Breast cancer (HER2, PR) & TG &
GIN & WSI; Supervised; Supervised; TCGA-BRCA~\cite{cancer2012comprehensive} (2 classes); Graph-level. Status of Human epidermal growth factor receptor 2 (HER2) and Progesterone receptor (PR). \\
Ayg{\"u}ne{\c{s}} et al. (2020)~\cite{aygunecs2020graph} & Classification &  Breast cancer & PG &
GCN & WSI; Supervised; Weakly-supervised; Department of Pathology at Hacettepe University (private) (4 classes). ROI-level classification. \\
Ye et al. (2019)~\cite{ye2019improving} & Classification & Breast cancer & PG &
GCN & WSI; Supervised; Supervised; BACH~\cite{aresta2019bach} (4 classes); Graph construction based on the ROI segmentation map. \\
Zhao et al. (2020)~\cite{zhao2020predicting} & Classification & Colorectal cancer & PG &
ChebNet, SAGPool & WSI; Self-Supervised; Weakly-supervised;  TCGA-COAD~\cite{kandoth2013mutational} (2 classes); Multiple instance learning. Graph-level output. \\
Raju et al. (2020)~\cite{raju2020graph} & Classification & Colorectal cancer & TG &
Adaptive GraphSage + Attention & WSI; Self-Supervised; Weakly-supervised; MCO~\cite{ward2015molecular} (4 classes); Multiple instance learning. Cluster embedding (Siamese architecture); Tumor node metastasis staging. \\
Ding et al. (2020)~\cite{ding2020feature} & Classification &  Colorectal cancer & PG &
Spatial-GCN (FENet) & WSI; Supervised; Supervised; TCGA-COAD and TCGA-READ~\cite{kirk2016radiology} (2 classes); Genetic mutational prediction. \\
Adnan et al. (2020)~\cite{adnan2020representation} & Classification & Lung cancer & PG &
ChebNet, GraphSAGE + Global attention pooling & WSI; Supervised; Supervised; TCGA-LUSC~\cite{tomczak2015cancer} (2 classes),  MUSK1~\cite{dua2017uci}; Adjacency learning layer. Multiple instance learning. \\
Zheng et al. (2019)~\cite{zheng2019encoding} & Retrieval & Lung cancer & PG &
GNN, DiffPool (GNN-Hash) & WSI; Supervised; Similarity (Hamming distance); ACDC-LungHP~\cite{tomczak2015cancer}; Hashing methods and binary encoding. Histopathological image retrieval.  \\
Li et al. (2018)~\cite{li2018graph} & Classification & Lung cancer & PG &
ChebNet + Attention & WSI; Self-Supervised; Supervised; TCGA-LUSC~\cite{tomczak2015cancer} (2 classes), NLST~\cite{kramer2011lung} (2 classes); Survival prediction. \\
Wu et al. (2019)~\cite{wu2019weakly} & Classification & Skin cancer & PG &
GCN & WSI; Supervised; Weakly- and Semi-supervised; BCC data collected from 2 different hospitals (private) (4 classes). \\
Anklin et al. (2021)~\cite{anklin2021learning} & Segmentation / Classification & Prostate cancer & TG &
GIN (SegGini) + GraphGrad-CAM & TMA, WSI; Supervised; Weakly-supervised; UZH prostate TMAs~\cite{zhong2017curated} (4 classes), SICAPv2~\cite{silva2020going} (4 classes); Gleason grade, Post-hoc interpretability. \\ 
\midrule
%
Pati et al. (2021)~\cite{pati2021hierarchical}  & Classification & Breast cancer & CG, TG, HR &
GIN-PNA (HACT-Net) + GraphGrad-CAM & WSI; Supervised; Supervised; BRACS~\cite{pati2020hact} (7 classes), BACH~\cite{aresta2019bach} (4 classes); Cell-to-Tissue Hierarchies. \\
Pati et al. (2020)~\cite{pati2020hact} & Classification & Breast cancer & CG, TG, HR &
GIN (HACT-Net) & WSI; Supervised; Supervised; BRACS~\cite{pati2020hact} (5 classes); Cell-to-Tissue Hierarchies. \newline\\ 
Zhang and Li (2020)~\cite{zhang2020ms} & Classification & Breast cancer & PG, HR &
MS-GWNN & WSI; Supervised; Supervised; BACH~\cite{aresta2019bach} (4 classes), BreakHis~\cite{spanhol2015dataset} (2 classes); Multi-scale graph feature learning (node-level and graph-level prediction). \\
Levy et al. (2021)~\cite{levy2020topological} & Regression & Colorectal cancer / lymphoma & PG, HR &
GAT, TDA + Graph Mapper & WSI; Supervised; Supervised; Dartmouth Hitchcock Medical Center (private): colon (9 classes), lymph (4 classes); Hierarchical representation. Tumor invasion score and staging. \\
%
\midrule
Shi et al. (2020)~\cite{shi2020cervical}  & Classification & Cervical cancer & CCG &
Fusion CNN-GCN & RGB; Supervised; Semi-supervised; SIPaKMed~\cite{plissiti2018sipakmed} (5 classes), Motic~\cite{Shi2019GraphCN} (7 classes); Population analyis of isolated cell images. \\
Shi et al. (2019)~\cite{Shi2019GraphCN}  & Classification & Cervical cancer & CCG &
Fusion CNN-GCN & RGB; Supervised; Supervised; SIPaKMed~\cite{plissiti2018sipakmed} (5 classes), Motic~\cite{Shi2019GraphCN} (7 classes); Population analyis of isolated cell images. \\
Chen et al. (2020)~\cite{chen2020pathomic} & Classification & Renal Cancer & CG &
GraphSAGE, SAGPool + Attention & Fusion: WSI+Genome; Self-Supervised; Self-Supervised; TCGA-GBMLGG, TCGA-KIRC~\cite{tomczak2015cancer}; Survival outcome, Integrated gradient method.  \\
\bottomrule
\multicolumn{6}{p{500pt}}
{ 
Graph representation: Cell-Graph (CG); Patch-Graph (PG); Tissue-Graph (TG); Hierarchical Representation (HR); Cluster-Centroids-Graph (CCG) 
}
\end{tabular}}
\vspace{-4pt}
\end{table*}

%
%
\vspace{-3pt}
\section{Applications of graph deep learning \\ in digital pathology}
\label{sec:sec3}%

The case studies presented in this section are organised according to the methodology adopted for the graph representation and the clinical application. The graph model, training paradigm, and datasets used in all applications are detailed in Table~\ref{table:pathology}. 
Rather than providing an exhaustive review of the literature, we present prominent highlights concerning the pre-processing, graph construction and graph models adopted, and their benefits in addressing various pathology tasks.

With the development of TMAs and WSI scanning techniques, as well as access to massive digital datasets of tissue images, deep learning methods for tumor localization, survival prediction and cancer recurrence prediction have made substantial progress~\cite{bera2019artificial}. 
Both the spatial arrangement of cells of various types (macro features), and the details of specific cells (micro features) are important for detecting and characterizing cancers. Thus, a valuable representation of histopathology data must capture micro features and macro spatial relationships.
Graphs are powerful representational data structures, and have attracted significant attention in analysis of histopathological images~\cite{sharma2016cell} due to their ability to represent tissue architectures. The paradigm change from pixel-based to entity-based research has the potential to improve deep learning techniques' interpretability in digital pathology, which is relevant for diagnostics.

\vspace{-6pt}
\subsection{Cell-graph representation}

Most of these works follow a similar framework where a cell-graphs is introduced using cells as the entities to capture the cell micro-environment. The image is converted into a graph representation with the locations of identified cells serving as graph vertices and edges constructed depending on spatial distance. Cell-level features are extracted as the initial node embedding. The cell-graph is fed to a GCN to perform image-wise classification.

\subsubsection{Breast cancer}
Breast cancer is the most commonly diagnosed cancer and registers the highest number of cancer deaths among women. A majority of breast lesions are diagnosed along a spectrum of cancer classes that ranges from benign to invasive.
Cancer diagnosis and the detection of breast cancer is one of the most common applications of machine learning and computer vision within digital pathology analysis. CNNs have been used for various digital pathology tasks in breast cancer diagnosis such as nucleus segmentation and classification, and tumor detection and staging. However, these patch-wise approaches do not explicitly capture the inter-nuclear relationships and limit access to global information.

Anand et al.~\cite{anand2020histographs} proposed the use of GCNs to classify WSIs represented by graphs of their constituent cells. 
Micro-level features (nuclear morphology) were incorporated as vertex features using local image descriptors, while macro-level features (gland formation) were included as edge attributes based on a mapping of Euclidean distances between nearby nuclei.
The vertex features are represented by the average RGB intensity, morphological features and learned features extracted from a pre-trained CNN applied to a window around the nuclei centroid. Finally, each tissue image is classified by giving its cell-graph as an input to the GCN which is trained in a supervised manner. The authors adopted a spatial GCN known as robust spatial filtering (RSF)~\cite{such2017robust}, which can take heterogeneous graphs as input. This framework is depicted in Fig.~\ref{fig:Fig3}.
The authors demonstrate competitive performance compared to conventional patch-based CNN approaches to classify patients into cancerous or non-cancerous groups using the Breast Cancer Histology Challenge (BACH) dataset~\cite{aresta2019bach}.

Sureka et al.~\cite{sureka2020visualization} modeled histology tissue as a graph of nuclei and employed the RSF with a GCN~\cite{such2017robust} with attention mechanisms and node occlusion to highlight the relative cell contributions in the image, which fits the mental model used by pathologists. 
In the first approach, the authors occluded nuclei clusters to assess the drop in the probability of the correct class, while also including a method based on~\cite{gadiya2018some} to learn enhanced vertex and edge features. 
In a second approach, an attention layer is introduced before the first pooling operation for visualization of important nuclei for the binary classification of breast cancer on the BACH dataset and Gleason grade classification on a prostate cancer~\cite{arvaniti2018automated} dataset. 

\begin{figure}[!t]
\centering
\includegraphics[width=1\linewidth]{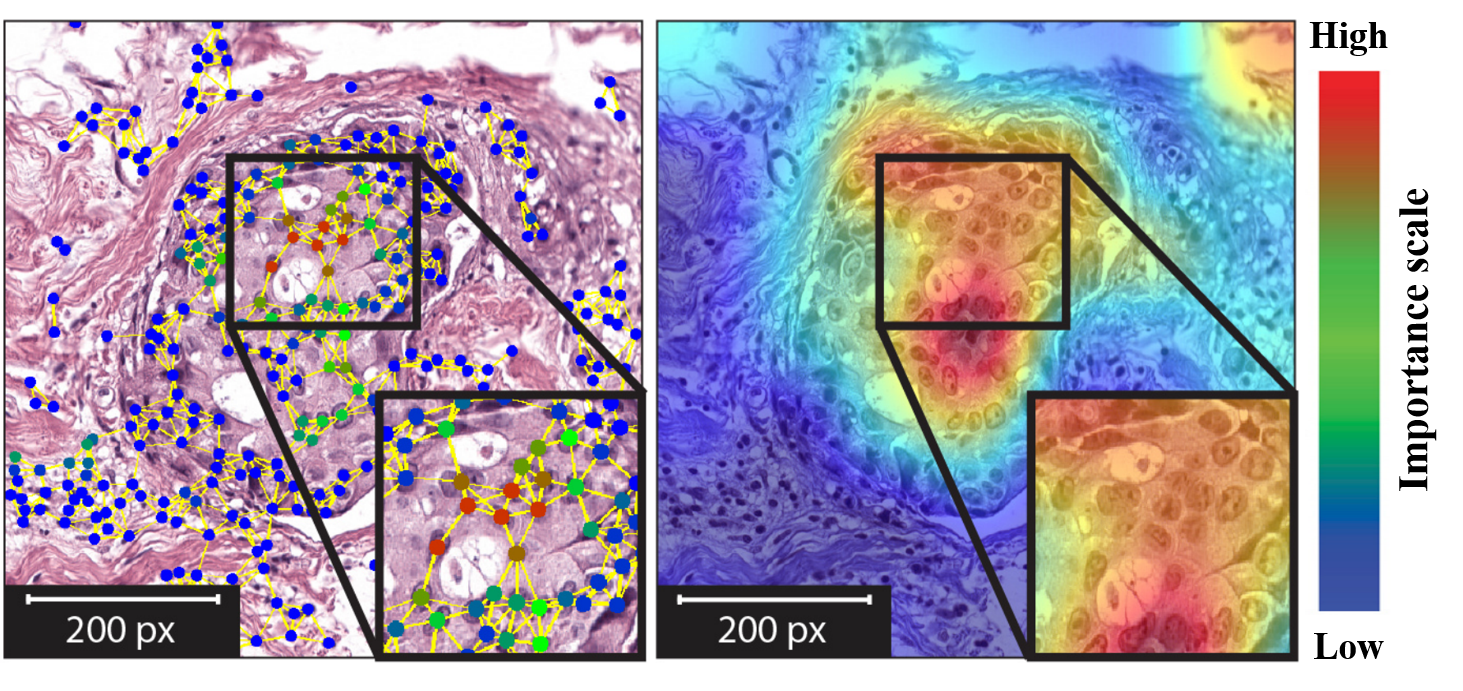}
\caption{
For a ductal carcinoma, examples of explanations given by graph-based (Left) and pixel-based (Right) explainability algorithms. Recreated from~\cite{jaume2020quantifying}.
}
\label{fig:Fig4}
\vspace{-10pt}
\end{figure}

Several explainers have been applied in digital pathology, inspired by explainability techniques for CNN model predictions on images. However, pixel-level explanations fail to encode tumor macro-environment information, and result in ill-defined visual heatmaps of important locations as illustrated in Fig.~\ref{fig:Fig4}. Thus, graph representations are relevant for both diagnostics and interpretation. Generating intuitive explanations for pathologists is critical to quantify the quality of the explanation.
To address this, Jaume et al.~\cite{jaume2020quantifying} introduced a framework using entity-based graph analysis to provide pathologically-understandable concepts (\textit{i.e.} to make the graph decisions understandable to pathologists).
The authors proposed a set of quantitative metrics based on pathologically measurable cellular properties to characterize explainability techniques in cell-graph representations for breast cancer sub-typing. 

In~\cite{jaume2020quantifying}, the authors first transform the histology image into a cell-graph, and a GIN model is used to map the corresponding class level. Then, a post-hoc graph explainer generates an explanation per entity graph. Finally, the proposed metrics are used to assess explanation quality in identifying the nuclei driving the prediction (nuclei importance maps). 
Four graph explainers were considered in this analysis: 
GNNExplainer~\cite{ying2019gnnexplainer}, GraphGrad-CAM~\cite{selvaraju2017grad},  GraphGrad-CAM++~\cite{chattopadhay2018grad}, and GraphLRP~\cite{schwarzenberg2019layerwise}.
The results on the Breast Carcinoma Subtyping (BRACS) dataset~\cite{pati2020hact} confirm that GraphGrad-CAM++ produces the best overall agreement with pathologists. The proposed metrics, which include domain-specific user-understandable terminology, could be useful for quantitative evaluation of graph explainability.

Jaume et al.~\cite{jaume2020towards} focused on the analysis of cells and cellular interactions in breast cancer sub-typing classification, and introduced an instance-level post-hoc graph-pruning explainer to identify decisive cells and interactions from the input graph in the BRACS dataset~\cite{pati2020hact}.
To create the cell-graph, nuclei are detected with segmentation algorithms and hand-crafted features including shape, texture and color attributes are extracted to represent each nucleus. The cell-graph topology uses the KNN algorithm and is based on the assumption that that spatially close cells encode biological relationships and, as a result, should create an edge. 
The cell-graph is processed by a GIN model, followed by a MLP to predict the cancer stages.

\begin{figure}[!t]
\centering
\includegraphics[width=0.95\linewidth]{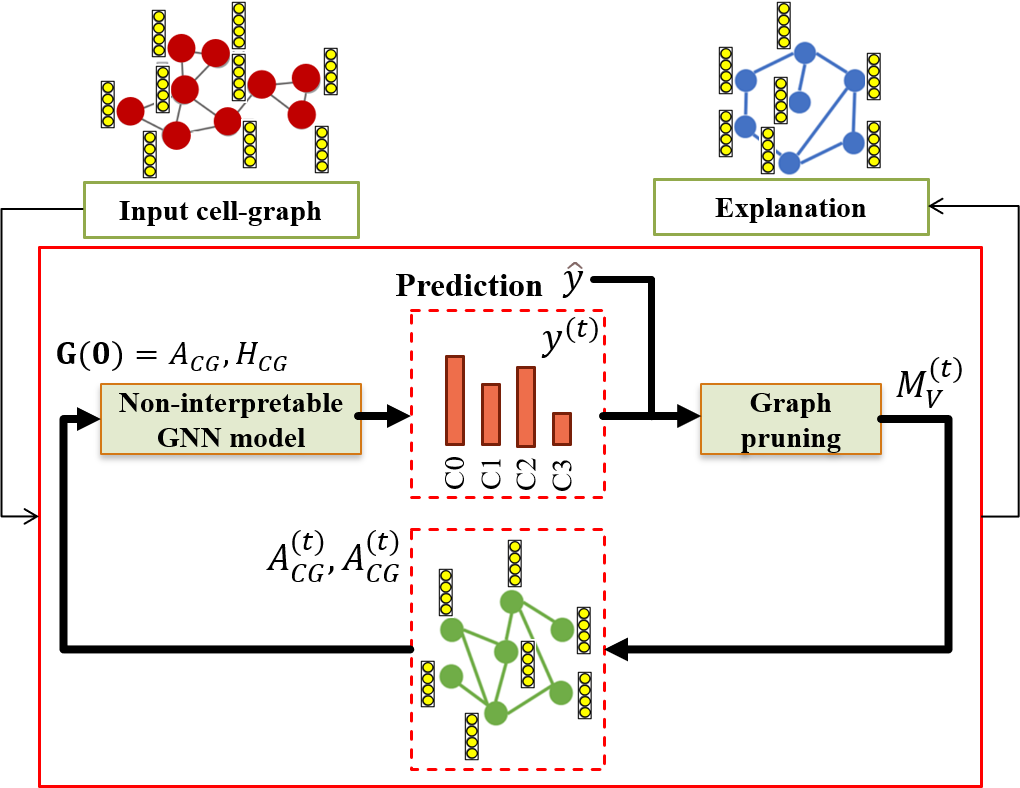}
\caption{
Cell-graph explainer (CGExplainer): a customized post-hoc graph explainer based on graph pruning optimization. Recreated from~\cite{jaume2020towards}.
}
\label{fig:Fig5}
\vspace{-10pt}
\end{figure}

Jaume et al.~\cite{jaume2020towards} designed a cell-graph explainer (CGExplainer), based on the GNNExplainer, to remove redundant and uninformative graph components, and the resulting sub-graph will be responsible for class-specific patterns that will aid disease comprehension. 
This module aims to learn a mask at the node-level that activates or deactivates parts of the graph. Fig.~\ref{fig:Fig5} provides an overview of the explainer module. The proposed explainer was shown to prune a substantial percentage of nodes and edges to extract valuable information while retaining prediction accuracy (\textit{e.g.} the explanations retain relevant tumor epithelial nuclei for cancer diagnosis).

\subsubsection{Colorectal cancer}
Colorectal cancer (CRC) grading is a critical task since it plays a key role in determining the appropriate follow-up treatment, and is also indicative of overall patient outcome. The grade of a cancer is determined, for example, by assessing the degree of glandular formation in the tumour. Nevertheless, automatic CNN-based methods for grading CRC typically use image patches which fail to include information on the micro-architecture of the entire tissue sample, and do not capture correspondence between the tissue morphology and glandular structure. 
To model nuclear features along with their cellular interactions, Zhou et al.~\cite{zhou2019cgc} proposed a cell-graph model for grading CRC, in which each node is represented by a nucleus within the original image, and cellular interactions are captured as graph edges based on node similarity.
A nuclear instance segmentation model is used to detect the nucleus and to extract accurate node features including nucleus shape and appearance features. Spatial features such as centroid coordinates, nuclei intensity and dissimilarity extracted from the grey level co-occurrence matrix were used as descriptors for predicting the grade of cancer. To reduce the number of nodes and edges based on the relative inter-node distance, an additional sampling strategy was used.

To conduct the graph-level classification, the authors in~\cite{zhou2019cgc} proposed the Adaptive GraphSAGE model, which is inspired by GraphSAGE~\cite{hamilton2017inductive} and JK-Net~\cite{xu2018representation}, to obtain multi-level features (\textit{i.e.} capturing the gland structure at various scales). 
To achieve multi-scale feature fusion, Adaptive GraphSAGE employs an attention technique which allows the network to adaptively generate an effective node representation. 

A graph clustering operation, which can be considered as an extension of DiffPool{~\cite{ying2018hierarchical}}, is used to group cells according to their appearance and tissue type, and to extract more abstract features for hierarchical representation. However, since the tissue hierarchy is inaccessible via this approach, the representation does not include high-level tissue features.
Based on the degree of gland differentiation, the graph model categorises each image as normal, low-grade, or high-grade. In comparison with a traditional CNN, the proposed model achieves better accuracy by incorporating both nuclear and graph-level features.

\begin{figure*}[!t]
\centering
\includegraphics[width=1\linewidth]{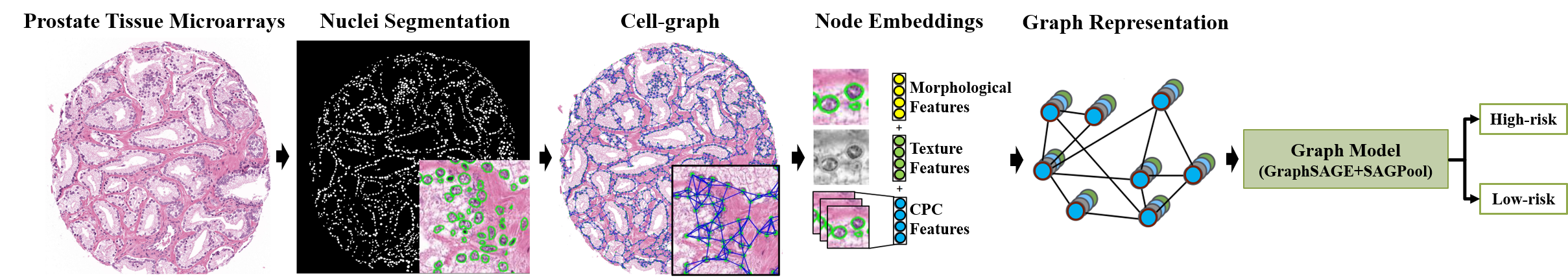}
\vspace{-10pt}
\caption{
The nuclei that have been detected are segmented, and a graph is constructed using the centroid of each nuclei. For each node, morphological, texture and contrastive predictive coding features are extracted, and GCNs are used as the graph representation. Recreated from~\cite{wang2020weakly}.
}
\label{fig:Fig6}
\vspace{-8pt}
\end{figure*}

Dysplasia of intestinal glands is especially important in pT1 colorectal cancer, the earliest stage of invasive colorectal cancer. 
Studer et al.~\cite{studer2019graph} introduced the pT1 Gland graph (pT1-GG) dataset that consists of cell-graphs of healthy and dysplastic intestinal glands. In this work, the authors established a baseline for gland classification using labelled cell-graphs and the graph edit distance (GED), which is an error-tolerant measurement of similarity between two graphs. This technique is an improved version of the bipartite graph-matching method (BP2)~\cite{fischer2017improved} combined with a KNN algorithm to perform classification.

Later, the same authors investigated different graph-based architectures~\cite{studer2021classification} to classify healthy gland tissue and dysplastic glandular areas on the pT1-GG dataset.
The GNN architectures evaluated for cell-graph classification are GCN~\cite{kipf2017semi}, GraphSAGE~\cite{hamilton2017inductive}, GAT~\cite{velivckovic2017graph}, GIN~\cite{xu2018powerful}, EGNN~\cite{gong2019exploiting} and a 1-dimensional GNN~\cite{morris2019weisfeiler}. All models are trained using three graph convolution layers where GraphSAGE and GCN are also trained with jumping knowledge (JK)~\cite{xu2018representation} to allow for an adaptive neighborhood range by aggregating representations across different layers. A concatenation of global sum-pooling, global mean-pooling and global max-pooling is used to get the graph-level output, followed by a MLP to classify an input graph.
The results demonstrated that graph-based deep learning methods outperformed classical graph-based and CNN-based methods. 
It should be emphasised, however, that each node is only linked to its two spatially closest neighbors, resulting in very restricted information sharing during message passing.

\subsubsection{Prostate cancer}
The commonly used Gleason score, which is based on the architectural pattern of tumor tissues and the distribution of glands, determines the aggressiveness of prostate cancer. CNNs have been used for histology image classification including Gleason score assignment, but CNNs are unable to capture the dense spatial relationships between cells and require detailed pixel level annotations for training.

To analyse the spatial distribution of the glands in prostate TMAs, Wang et al.~\cite{wang2020weakly} proposed a weakly-supervised approach for grade classification and to stratify low and high-risk cases (Gleason score $<6$ is normal tissue; Gleason score $\geq 6$ is abnormal tissue or high-risk).
The authors segmented the nuclei and construct a cell-graph for each image with nuclei as the nodes, and the distance between neighboring nuclei as the edges, as illustrated in Fig.~\ref{fig:Fig6}.
Using prostate TMAs with only image-level labels rather than pixel-level labels, a GCN is used to identify high-risk patients via a self-supervised technique known as contrastive predictive coding (CPC)~\cite{oord2018representation}.
Features for each node are generated by extracting morphological (area, roundness) and texture features (dissimilarity, homogeneity) as well as features from CPC-based learning.
A GraphSAGE convolution and a self-attention graph pooling (SAGPool)~\cite{lee2019self} are applied to the graph representation to learn from the global distribution of cell nuclei, cell morphology and spatial features. The proposed method can calculate attention scores, focus on the more significant node attributes, and aggregate information at different levels.

\vspace{-6pt}
\subsection{Patch-graphs and Tissue-graphs representations}  

The majority of the following works transform pathological images into patch-graphs, where nodes are important patches, and edges encode the intrinsic relationships between these patches.
These patches are sampled using methods such as color-based, cell density or attention mechanisms. Then, CNNs are used to extract features from these patches to generate a feature vector for the node embedding of the graph representation. Given the constructed graph, a graph deep learning model is used to conduct node or graph classification.
It is important to make the distinction between tissue-graphs, which are biologically-defined and capture relevant morphological regions; while patch-graphs connect patches of interest, where each patch can contain multiple biological entities, with each other.

\begin{figure*}[!t]
\centering
\includegraphics[width=0.9\linewidth]{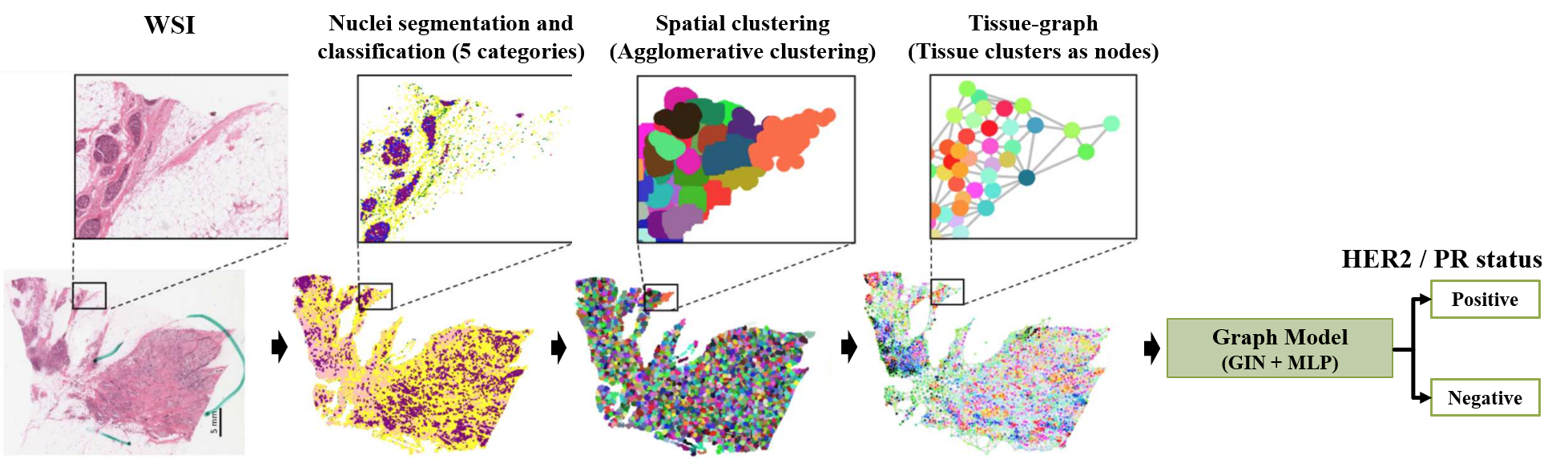}
\caption{
Slide Graph Model which constructs a graph from the nuclei level to the entire WSI-level. The main steps are as follows: segmenting and classification of nuclei; clustering; constructing the graph and graph classification. Recreated from~\cite{lu2020capturing}.
}
\label{fig:Fig8}
\vspace{-6pt}
\end{figure*}

\subsubsection{Breast cancer} 
Multi-class classification of arbitrarily sized ROIs is an important problem that serves as a necessary step in the diagnostic process for breast cancer.
Ayg{\"u}ne{\c{s}} et al.~\cite{aygunecs2020graph} proposed to incorporate local context through a graph-based ROI representation over a variable number of patches (nodes) and their spatial proximity relationships (edges).
A CNN is used to extract a feature vector for each node represented by fixed-sized patches of the ROI.
Then, to propagate information across patches and incorporate local contextual information, two consecutive GCNs are used, which also aggregate the patch representation to classify the whole ROI into a diagnostic class. 
The classification is conducted in a weakly-supervised manner over the patches and ROI-level annotations, without having access to patch-level labels.
Results on a private data collected from the Department of Pathology at Hacettepe University outperformed CNN-based models that incorporated majority-voting, learned-fusion and base-penultimate methods.

Some traditional CNN-based models have proposed to jointly segment a ROI of an image and classify WSIs and that enabled the classifier to better predict the image class~\cite{mehta2018net}.
Ye et al.~\cite{ye2019improving} captured the topological structure of a ROI image through a GCN where a graph is constructed with segmentation masks of image patches that contain high levels of semantic information.
The segmentation mask for each image patch is obtained using an encoder-decoder semantic segmentation framework where each pixel is classified as one of the four classes of tissue samples (normal, benign, in situ, and invasive) of the BACH~\cite{aresta2019bach} dataset.
The combined segmentation masks of the image patches yield the total ROI segmentation mask. The area ratio of each lesion is calculated as the value of the unit node in each picture patch. Then, a graph is constructed to capture the spatial dependencies using the features of the image patch segmentation masks. Finally, the ROI image is classified based on the features learned by the GCNs. 

One limitation of previous works is that they construct graphs using small patches of the WSI. Lu et al.~\cite{lu2020capturing} overcome this challenge by introducing a pipeline to construct a graph from the entire WSI using the nuclei level information, including geometry and cellular organization in tissue slides (termed the histology landscape). 
After building the graph, the authors used a GIN model to predict the positive or negative human epidermal growth factor receptor 2 (HER2), and the progesterone receptor (PR), which are two valuable biomarkers for breast cancer prognosis.

The proposed method in~\cite{lu2020capturing} consists of four steps as illustrated in Fig.~\ref{fig:Fig8}. 
This work first used Hover-Net~\cite{graham2019hover} to simultaneously segment and classify the individual nuclei and extract their features. Then, agglomerative clustering~\cite{mullner2011modern} is used to group spatially neighboring nuclei into clusters which results in reduced computational cost for downstream analysis. Using these clusters, a graph is generated by assigning the tissue clusters to nodes and the edges of the graph encode the cellular topology of the WSI. Lastly, the graph generated from the entire WSI is used as an input to a GCN to predict HER2 or PR status at the WSI-level.
The performance of this method is evaluated on the hematoxylin and eosin (H\&E) stained WSI images from the TCGA-BRCA~\cite{cancer2012comprehensive} dataset, which consist of 608 HER2 negative and 101 HER2 positive, and 452 PR positive and 256 PR negative samples.

Content-based histopathological image retrieval has also been investigated for decision support in digital pathology. This system scans a pre-existing WSI database for regions that the pathologist is interested in and returns related regions to the pathologists for comparison. These methods can provide valuable information including diagnosis reports from experts for similar regions. Retrieval methods can also be used for classification by considering the most likely diagnosis~\cite{zheng2018histopathological}. However the amount of manually labelled training data limits their power. 
Ozen et al.~\cite{ozen2021self} suggested a generic method that combines GNNs with a self-supervised training method that employs a contrastive loss function without requiring labeled data. 
In this framework, fixed-size patches and their spatial proximity relations are represented by undirected graphs. 
The simple framework for constrastive learning of visual representation (SimCLR)~\cite{chen2020simple} is adopted for learning representations of ROIs.
Using the contrastive loss, the GNN encoder and MLP projection head are trained to maximise the agreement between the representations. A GCN followed by a DiffPool operation is selected as the model configuration. 

For content-based retrieval tasks, this GNN is trained in a self-supervised setting and is used to extract ROI representations where the Euclidean distance between the extracted representations is used to determine how similar two ROIs are. 
Quantitative results demonstrated that contrastive learning can improve the quality of learned representations, and despite not utilizing class labels could outperforming supervised classification methods.

\begin{figure*}[!t]
\centering
\includegraphics[width=1\linewidth]{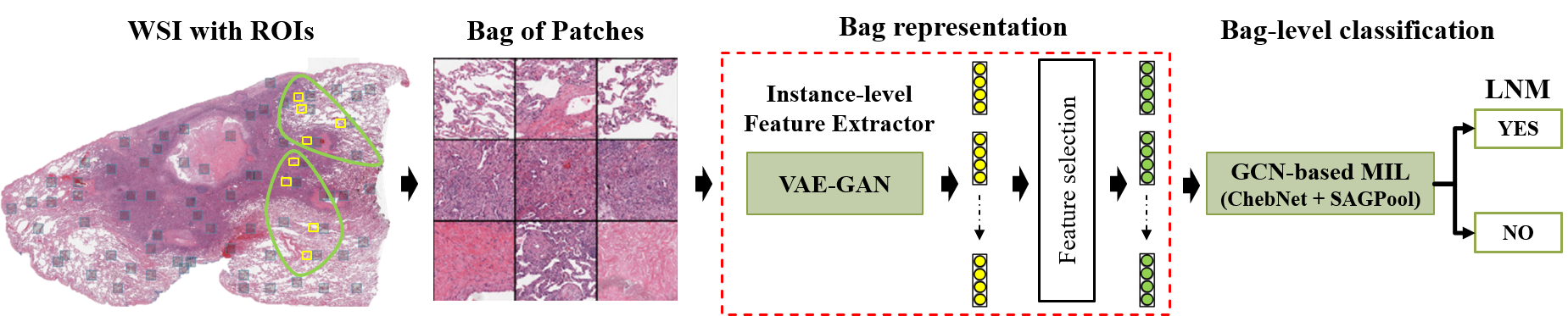}
\caption{
Representation of a GCN-based MIL method. Once the bag of patches are extracted, instance-level feature extraction and selection is conducted followed by a bag-level classification. Recreated from~\cite{zhao2020predicting}.
}
\label{fig:Fig9}
\vspace{-3pt}
\end{figure*}

\begin{figure*}[!t]
\centering
\includegraphics[width=0.9\linewidth]{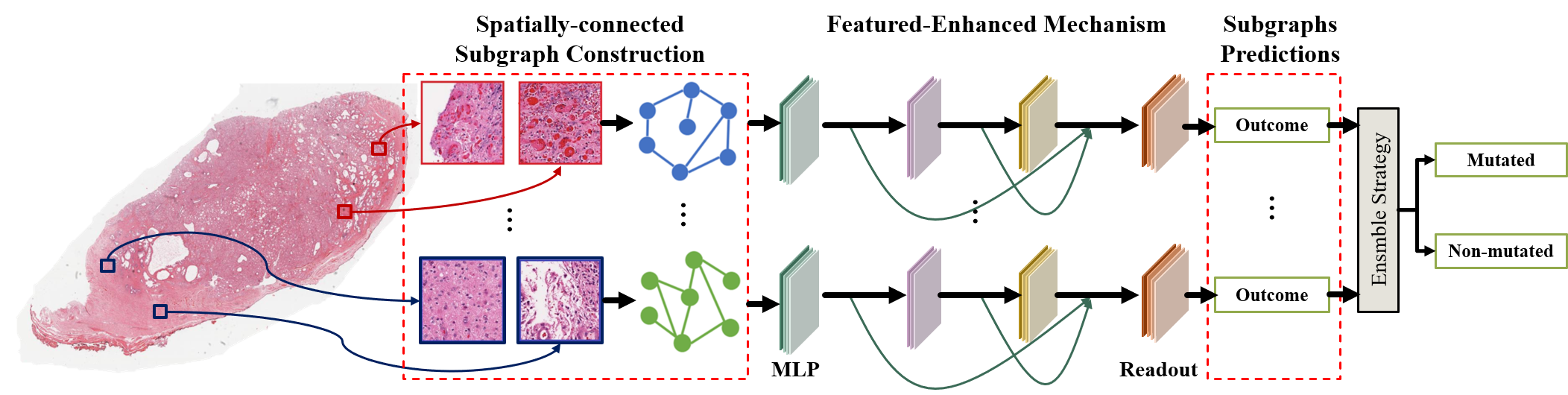}
\vspace{-2pt}
\caption{
The proposed FENet architecture. For each WSI, patches are randomly selected. Each patch corresponds to a node in each non-isomorphic subgraph where a CNN is used to extract node attributes. A feature-enhanced mechanism is adopted to consider all topological structural information. An ensemble approach used majority voting to aggregate all subgraphs' prediction outcomes. Recreated from~\cite{ding2020feature}.
}
\label{fig:Fig10}
\vspace{-6pt}
\end{figure*}

\subsubsection{Colorectal cancer}

Although CNN-based approaches have practical merits when identifying important patches for predicting CRC, they do not take into account the spatial relationships between patches, which is important for determining the stage of the tumor. The size and the relative location of the tumor in relation to other tissue partitions are used for tumor node metastasis staging estimation. Furthermore, traditional approaches require the presence of expert pathologists to annotate each WSI.
Weakly-supervised learning is an important and potentially viable solution to dealing with sparse annotations in medical imagery. Multiple instance learning (MIL) is well-suited to histology slide classification, as it is designed to operate on weakly-labeled data~\cite{srinidhi2020deep}.

Raju et al.~\cite{raju2020graph} considered the spatial relationship between tumor and other tissue partitions with a graph attention multi-instance learning framework to predict colorectal tumor node metastasis staging. Each graph with nodes representing different tissues serves as an instance, and the multiple instances for a WSI form a bag that aids in tumour stage prediction.

In~\cite{raju2020graph}, given a WSI, a texture autoencoder~\cite{zhang2017deep} is used to encode the texture from random sample patches. Then a cluster embedding network based on a Siamese architecture~\cite{ye2019unsupervised} is trained on a binary classification task to group similar texture features into multiple graphs.
Each WSI is divided into multiple graphs and each graph has features from all cluster labels.
The authors used a tissue wise annotated CRC dataset~\cite{gupta2019prediction} to assign cluster labels for similar image patches.
The authors consider the multiple graphs as multiple instances in a bag which are used to predict the tumor staging using an attention MIL method~\cite{ilse2018attention}. The authors adopted an Adaptive GraphSage~\cite{zhou2019cgc} approach with learnable attention weights to assign more importance to instances which contain more information towards predicting the tumor stage.
The authors demonstrated that graph attention multi-instance learning can perform better than a GCN on the Molecular and Cellular Oncology (MCO)~\cite{ward2015molecular} dataset.

Colorectal cancer lymph node metastasis (LNM) is a crucial factor in patient management and prognosis, and its identification suggests the need for dissection to avoid further spread.
Zhao et al.~\cite{zhao2020predicting} introduced a GCN-based multiple instance learning method combined with a feature selection strategy to predict LNM in the colon adenocarcinoma (COAD) cohort of the Cancer Genome Atlas (TCGA) project~\cite{kandoth2013mutational}.
Following the MIL approach, the training dataset is composed of bags where each bag contains a set of instances. The goal of this work is to teach a model to predict the bag label, where only the bag-level label is available.

The overall framework has three major components: instance-level feature extraction, instance-level feature selection, and bag-level classification, as illustrated in Fig.~\ref{fig:Fig9}. 
First, non-overlapping patches are extracted from a WSI which is represented as a bag of patches. Since instance labels are unavailable, the authors introduced a combination of a variational autoencoder (VAE)~\cite{kingma2013auto} and a generative adversarial network (GAN) for fine-tunning the encoder component as an instance-level feature extractor in a self-supervised manner. In this VAE-GAN model, the architecture of the network for the decoder of the VAE and generator of the GAN is the same network. \
Then, a feature selection component is incorporated to remove redundant and unhelpful features to alleviate the workload when generating the bag representation. The maximum mean discrepancy is used to evaluate the feature importance. Finally, the authors employed ChebNet~\cite{defferrard2016convolutional} followed by SAGPool~\cite{lee2019self} to generate the bag representation and perform the bag-level classification.
The authors demonstrated that the proposed model outperformed CNN-based and attention-based MIL models.

Colon adenoma and carcinoma may occur as a result of a series of histopathological changes due to key genetic alterations. Thus, the ability to predict genetic mutations is important for the diagnosis of colon cancer.
Ding et al.~\cite{ding2020feature} proposed a feature-enhanced graph network (FENet) using a spatial-GCNs, based on GIN, to predict gene mutations across all three key mutational prediction tasks (APC, KRAS, and TP53) that are associated with colon cancer evolution. In this approach, multiple spatial graphs are created using randomly selected image patches from each patient's WSI. 

The feature-enhanced mechanism aggregates features from neighboring patches and combines them as the central node representation to increase feature learning performance. 
The authors introduced GlobalAddPooling as a READOUT function to convert the node representation into a graph representation. The prediction outcome for each sub-graph is classified by fully-connected layers.
Finally, an ensemble strategy combines the prediction results of all sub-graphs to predict mutated and non-mutated classes. 
Fig.~\ref{fig:Fig10} illustrates the proposed FENet networks.
The authors demonstrated that the integration of multiple sub-graph outcomes in the proposed model leads to a significant improvement in prediction performance on the Cancer Genome Atlas Colon Adenocarcinoma dataset~\cite{kirk2016radiology}, outperforming graph-based baseline models such as ChebNet, GraphSAGE and GAT.

\begin{figure*}[!t]
\centering
\includegraphics[width=0.9\linewidth]{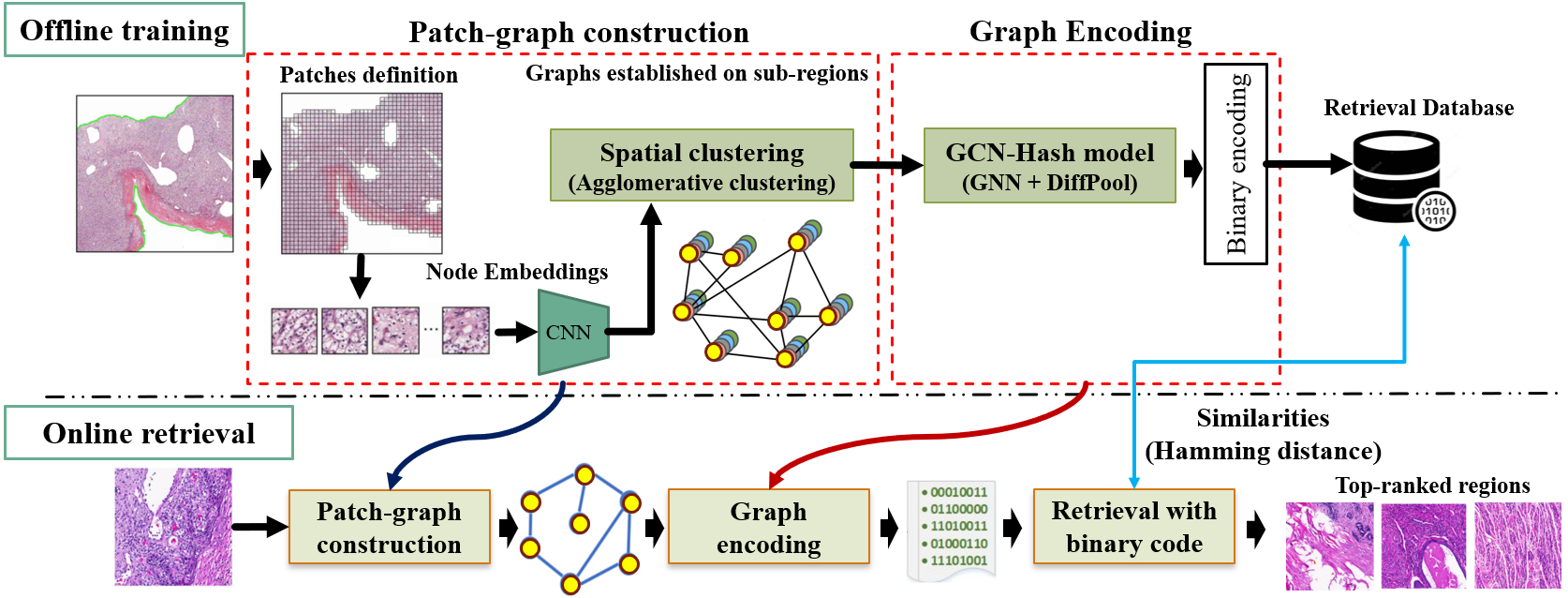}
\caption{
Representation of the retrieval framework. Patch-graphs are constructed based on spatial relationships and feature distances between patches, and are fed into the developed GNN-Hash model for graph encoding. 
When retrieving, the query region is converted into a patch-graph and a binary code for similarity comparison with samples in the database. Recreated from~\cite{zheng2019encoding}.
}
\label{fig:Fig12}
\vspace{-6pt}
\end{figure*}

\subsubsection{Lung cancer}
Lung adenocarcinoma and lung squamous cell carcinoma are the most common subtypes of lung cancer, and distinguishing between them requires a visual examination by an experienced pathologist.
Efficient mining of survival-related structural features on a WSI is a promising way to improve survival analysis. 
Li et al.~\cite{li2018graph} introduced a GCN-based survival prediction model that integrated local patch features with global topological structures (patch-graph) through spectral graph convolution operators (ChebNet) using the TCGA-LUSC~\cite{tomczak2015cancer} and NLST~\cite{kramer2011lung} datasets.
The model utilized a survival-specific graph trained under supervision using survival labels.
A parallel graph attention mechanism is used to learn attention node features to improve model robustness by reducing the randomness of patch sampling (\textit{i.e.} an adaptive patch selection by learning the importance of individual patches). 
This attention network is trained jointly with the prediction network. The authors demonstrated that topological features fine-tuned with survival-specific labels outperformed CNN-based models. 

Adnan et al.~\cite{adnan2020representation} explored the application of GNNs for MIL. The authors sampled important patches from a WSI and model them as a fully-connected graph where the graph is converted to a vector representation for classification. Each instance is treated as a node of the graph in order to learn end-to-end relationships between nodes.
In this approach, a DenseNet is used to extract features from all important patches sampled from a segmented tissue using color thresholds~\cite{kalra2020yottixel}. 
Then, an adjacency learning layer which uses global information about the patches is adopted to define the connections within nodes in an end-to-end manner. The adjacency matrix is calculated by an adjacency learning block using a series of dense layers and cross-correlation.
The constructed graph is passed through two types of graph models (ChebNet and GraphSAGE), followed by a graph pooling layer to get a single feature vector to compare the discrimination of sub-types of lung cancer on the TCGA{~\cite{tomczak2015cancer}} and MUSK1{~\cite{dua2017uci}} datasets. 
With the adopted global attention pooling{~\cite{li2015gated}} which uses a soft attention mechanism, it is possible to visualise the importance that the network places on each patch when making the prediction.
The pooled representation is fed to two fully connected dense layers to achieve the final classification between lung adenocarcinoma and lung squamous cell carcinoma. The proposed model outperformed CNN-based models that use attention-MIL.

\begin{figure*}[!t]
\centering
\includegraphics[width=0.9\linewidth]{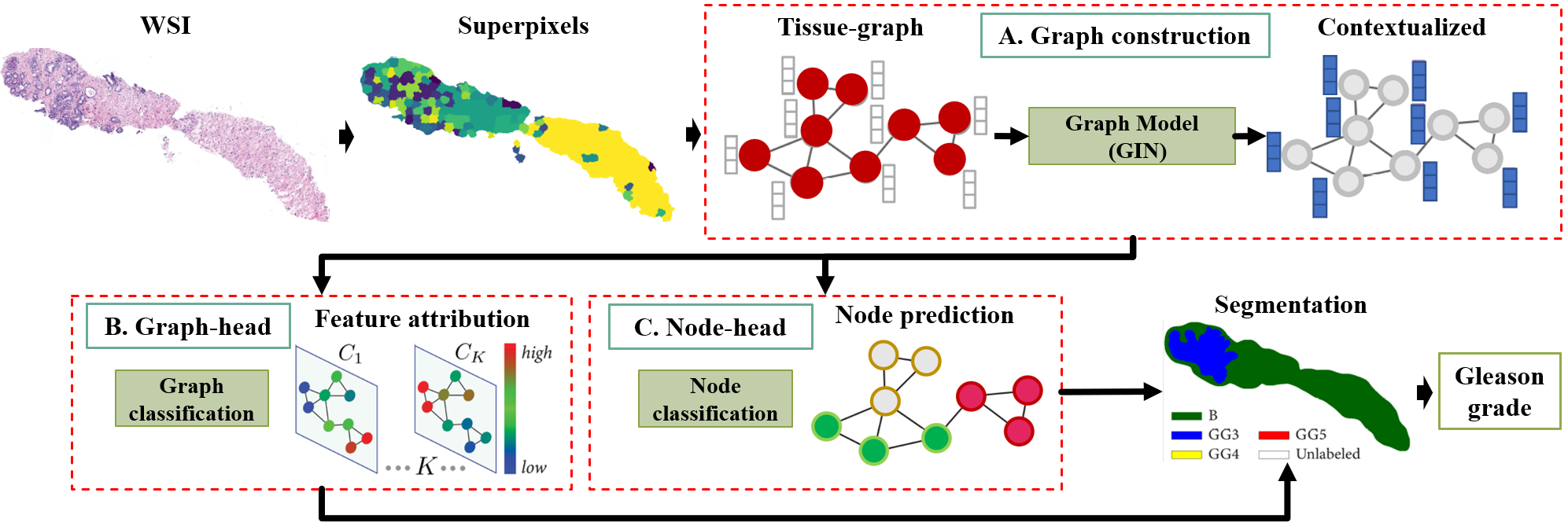}
\caption{
Representation of the proposed SegGini methodology.
a) Tissue graph construction with tissue superpixels as nodes, and edges computed using a region adjacency graph from the spatial connectivity of superpixels. A GNN is used to learn discriminative
node embeddings to perform semantic segmentation.
b) Graph-head: graph classification and feature atribution based on GraphGrad-CAM.
c) Node-head: node classification.
Recreated from~\cite{anklin2021learning}.
}
\label{fig:Fig13}
\vspace{-6pt}
\end{figure*}

As discussed previously, content-based image retrieval seeks to find images that have morphological characteristics that are most similar to a query image.
Binary encoding and hashing techniques have been successfully adopted to speed up the retrieval process in order to satisfy efficiency requirements~\cite{li2018large}. However, WSI are commonly divided into small patches to index WSIs for region-level retrieval. This process does not consider the contextual information from a broad region surrounding the nuclei and the adjacency relationships that exist for different types of biopsy.

Zheng et al.~\cite{zheng2019encoding} proposed a retrieval framework for a large-scale WSI database based on GNNs and hashing, which is illustrated in Fig.~\ref{fig:Fig12}.
Patch-graphs are first built in an offline stage based on patch spatial adjacency, and feature similarity extracted with a pre-trained CNN. Then, the patch-graphs are processed by a GNN-Hash model designed to use a graph encoding, and stored in the retrieval database. The GNN-Hash structure was created by stacking GNN modules and a DiffPool module~\cite{ying2018hierarchical}. The output of the hierarchical GNN-Hash is modified with a binary encoding layer in the final graph embedding layer. Finally, the relevant regions are retrieved and returned to pathologists after the region the pathologist queries is converted to a binary code. The similarities between the query code and those in the database are measured using Hamming distance.
Experiments to estimate the adjacency relationships between local regions in WSIs and the similarities with query regions were conducted using the lung cancer ACDC-LungHP~\cite{tomczak2015cancer} dataset.
The results demonstrated that the proposed retrieval model is scalable to different query region sizes and shapes, and returns tissue samples with similar content and structure.

\subsubsection{Skin cancer}
One of the most common types of skin cancer is basal cell carcinoma (BCC) which can look similar to open sores, red patches and shiny bumps. Several studies have demonstrated the ability to identify BCC from pathological images. 
Wu et al.~\cite{wu2019weakly} introduced a model that predicted BCC on WSI using a weakly- and semi-supervised formulation by combining prior knowledge from expert observations, and structural information between patches into a graph-based model. A sample of this prior knowledge is the fact that a dense patch with predictive cancer cells is more likely to have a cluster of cancer cells, and more patches with high cancer likelihoods increase the overall likelihood of an image being positive.

The framework consists of two modules, a GCN that propagates supervisory information over patches to learn patch-aware interpretabililty in the form of a probability score; and an aggregation function that connects patch-level and image-level predictions using prior knowledge. 
The proposed model makes full use of different levels of supervision, using a mix of weak supervision from image-level labels and  available pixel-wise segmentation labels as a semi-supervised signal.
By incorporating prior knowledge and structure information, both image-level classification and patch-level interpretation are significantly improved.

\subsubsection{Prostate cancer}
Pathologists must go above and beyond normal clinical demands and norms when precisely annotating image data. As a result, a semantic segmentation method should be able to learn from inexact, coarse, and image-level annotations without complex task-specific post-processing steps.
To this end, Anklin et al.~\cite{anklin2021learning} proposed a weakly-supervised semantic segmentation method based on graphs (SegGini) that incorporates both local and global inter-tissue-region relations to perform contextualized segmentation using inexact and incomplete labels.
The model is evaluated on the UZH (TMAs)~\cite{zhong2017curated} and SICAPv2 (WSI)~\cite{silva2020going} prostate cancer datasets for Gleason pattern segmentation and Gleason grade classification.
Fig.~\ref{fig:Fig13} depicts the proposed SegGini methodology.
A tissue-graph representation for an input histology image is constructed as proposed in~\cite{pati2020hact}, where the graph nodes depict tissue superpixels. As the rectangular patches can span multiple distinct structures, superpixels are used~\cite{bejnordi2015multi}. 
To characterize the nodes, morphological and spatial features are extracted, and the graph topology is computed with a region adjacency graph (RAG)~\cite{potjer1996region}, using the spatial connectivity of superpixels.

Given a tissue graph, a GIN model learns contextualized features from the tissue microenvironment and inter-tissue interactions to perform semantic segmentation, where the proposed SegGini model assigns a class label to each node.
The resulting node features are processed by a graph-head (image label), a node-head (node label), or both, based on the type of weak supervision.
The graph-head consists of a graph classification and a feature attribution technique. The authors employed GraphGrad-CAM to measure importance scores towards the classification of each class, where the node attribution maps determine the node labels.
Further, the authors in~\cite{anklin2021learning} found that the node-head simplifies image segmentation into classifying nodes where the node labels are extracted by assigning the most prevalent class within each node.
For inexact image label and incomplete scribbles, both heads are jointly trained to improve the individual classification tasks. 
The outcomes of the heads are used to segment Gleason patterns.
Finally, to identify image-level Gleason grades from the segmentation map, a classification approach~\cite{arvaniti2018automated} is used.
SegGini outperforms prior models such as HistoSegNet~\cite{chan2019histosegnet} in terms of per-class and average segmentation, as well as classification metrics. This model also provides comparable segmentation performance for both inexact and complete supervision; and can be applied to a variety of tissues, organs, and histology tasks.

\vspace{-6pt}
\subsection{Hierarchical graph representation (macro and micro architectures)}

In previous approaches, pathological images have been represented by cell-graphs, patch-graphs or tissue-graphs. However, cellular or tissue interactions alone are insufficient to fully represent pathological structures.
A cell-graph incorporates only the cellular morphology and topology, and discards tissue distribution information that is vital for appropriate representation of histopathological structures. A tissue-graph made up of a collection of tissue areas, on the other hand, is unable to portray the cell microenvironment.
Thus, to learn the intrinsic characteristics of cancerous tissue it is necessary to aggregate multilevel structural information, which seeks to replicate the tissue diagnostic process followed by a pathologist when analyzing images at different magnification levels.

\subsubsection{Breast cancer} 
Early detection of cancer can significantly reduce the mortality rate of breast cancer, where it is crucial to capture multi-scale contextual features in cancerous tissue. 
Combinations of CNNs have been used to encode multi-scale information in pathology images via multi-scale feature fusion, where scale is often associated with spatial location.

Zhang and Li~\cite{zhang2020ms} introduced a multi-scale graph wavelet neural network (MS-GWNN) that uses graph wavelets with different scaling parameters in parallel to obtain multilevel tissue structural information in a graph topology. The graph wavelet neural network (GWNN)~\cite{xu2019graph} replaces the graph convolution in a spectral GCN with the wavelet transform which has an excellent localization capability.
For breast cancer classification, the authors first transformed pathological images into graph structures where nodes are non-overlapping patches. Then, node classification is performed via a GWNN at different scales in parallel (node-level prediction). After that, multi-level node representations are incorporated to perform graph-level classification.
The results and the visualization of the learned node embeddings demonstrated the strong capacity of the model to encode different structural information on two public datasets: BACH~\cite{aresta2019bach} and BreakHis~\cite{spanhol2015dataset}. However, this approach is limited by the manual selection of the appropriate scaling parameter.

A hierarchy defined from the cells with learned pooling layers~\cite{zhou2019cgc} does not include high-level tissue features and approaches that concatenate cell-level and tissue-level information~\cite{chen2020pathomic} cannot leverage the hierarchy between the levels of the tissue representation.
To address these issues, Pati et al.~\cite{pati2020hact} proposed a hierarchical-cell-to-tissue (HACT) representation that utilizes both nuclei and tissue distribution properties for breast cancer subtype classification.
The HACT representation consists of a low-level cell-graph (CG) that captures the cellular morphology and topology; a tissue-graph (TG) at a high-level that captures the properties of the tissue sections as well as their spatial distribution; and the hierarchy between the cell-graph and the tissue-graph that captures the cells' relative distribution within the tissue.

\begin{figure}[!t]
\centering
\includegraphics[width=1\linewidth]{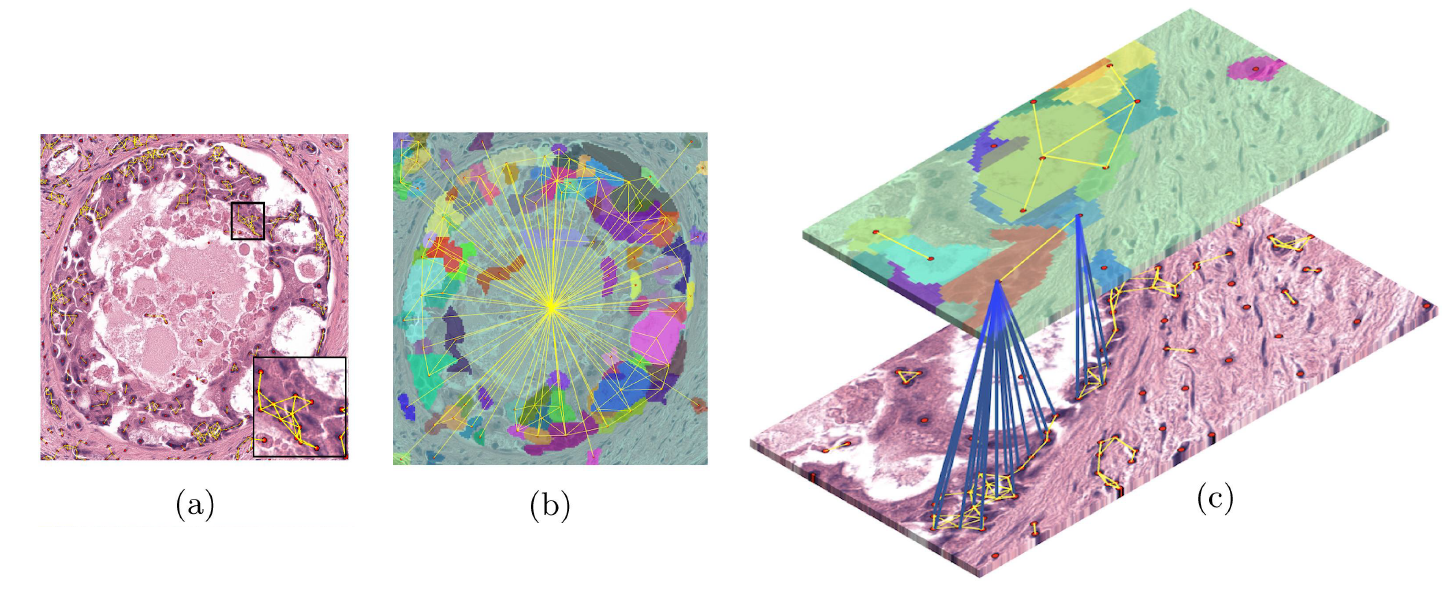}
\vspace{-6pt}
\caption{
Representation of a) CG, b) TG, and c) Hierarchical-cell-to-tissue. Image adapted from~\cite{pati2020hact}.
}
\label{fig:Fig14}
\vspace{-12pt}
\end{figure}

Fig.~\ref{fig:Fig14} illustrates samples of the CG, the TG and the hierarchical cell-to-tissue representation. 
To construct a CG, each node represents a cell and edges encode cellular interactions, where for each nucleus hand-crafted features such as shape, texture and spatial location are extracted. Then, a KNN algorithm is adopted to build the initial topology based on the assumption that a close cell should be connected and a distant cell should remain disconnected. The Euclidean distances between nuclei centroids in the image space are used to quantify cellular distances.
The TG is constructed by first identifying tissue regions (\textit{e.g.}, epithelium, stroma, lumen, necrosis) by detecting non-overlapping homogeneous superpixels of the tissue and iteratively merging neighboring superpixels that have similar colour attributes.
The TG topology is generated assuming that adjacent tissue parts should be connected by constructing a region adjacency graph~\cite{potjer1996region} with the spatial centroids of the superpixels.
The HACT representation, that jointly represents the low-level (CG) and high-level (TG) relationships, is processed with a hierarchical model (HACT-Net) that employs two GIN models~\cite{xu2018powerful}. The learned cell-node embeddings are combined with the corresponding tissue-node embeddings to predict the classes.

To demonstrate the hierarchical-learning, the authors introduce the BRACS dataset to classify five breast cancer subtypes: normal, benign, atypical, ductal carcinoma in situ, and invasive. The authors also evaluate the generalizability to unseen data by splitting the data at the WSI-level (two images from the same slide do not belong to different splits) different from previous approaches that split at the image-level~\cite{wang2020weakly,zhou2019cgc}.
The enriched multi-level HACT representation for classification outperformed CNN-based models and standalone cell-graph and tissue-graph models, confirming that for better structure-function mapping, the link between low-level and high-level information must be modelled at the local node level rather than at the graph level.

Later, Pati et al.~\cite{pati2021hierarchical} exploited hierarchical modeling for interpretability in digital pathology, aiming to map the tissue structure to tissue functionality. The authors adopt the hierarchical entity-graph representation of a tissue which is processed via a hierarchical GNN to learn the mapping from tissue compositions to respective tissue categories.
In this work, Pati et al.~\cite{pati2021hierarchical} improved the HACT representation and the HACT-Net model. HACT-Net is modeled using principal neighborhood aggregation (PNA)~\cite{corso2020principal} layers, which use a combination of aggregators to replace the sum operation in GIN and adopt degree-scalers to amplify or dampen neighboring aggregated messages based on the degree of a node.
Graph normalization followed by batch normalization is incorporated after each PNA layer~\cite{dwivedi2020benchmarking}, which aids the network in learning discriminative topological patterns when the number of nodes within a class varies dramatically.
To further assess the quality of the methodology, a comparison with independent pathologists is conducted. Three board-certified pathologists were recruited to annotate the BRACS test set without having access to the respective WSIs. The results indicate that the model outperforms the domain experts in the 7-class classification task.
The authors employed the GraphGrad-CAM to highlight the nuclei and tissue region nodes to show what the HACT-Net focuses on while classifying the tumor regions-of-interest.

\begin{figure}[!t]
\centering
\includegraphics[width=1\linewidth]{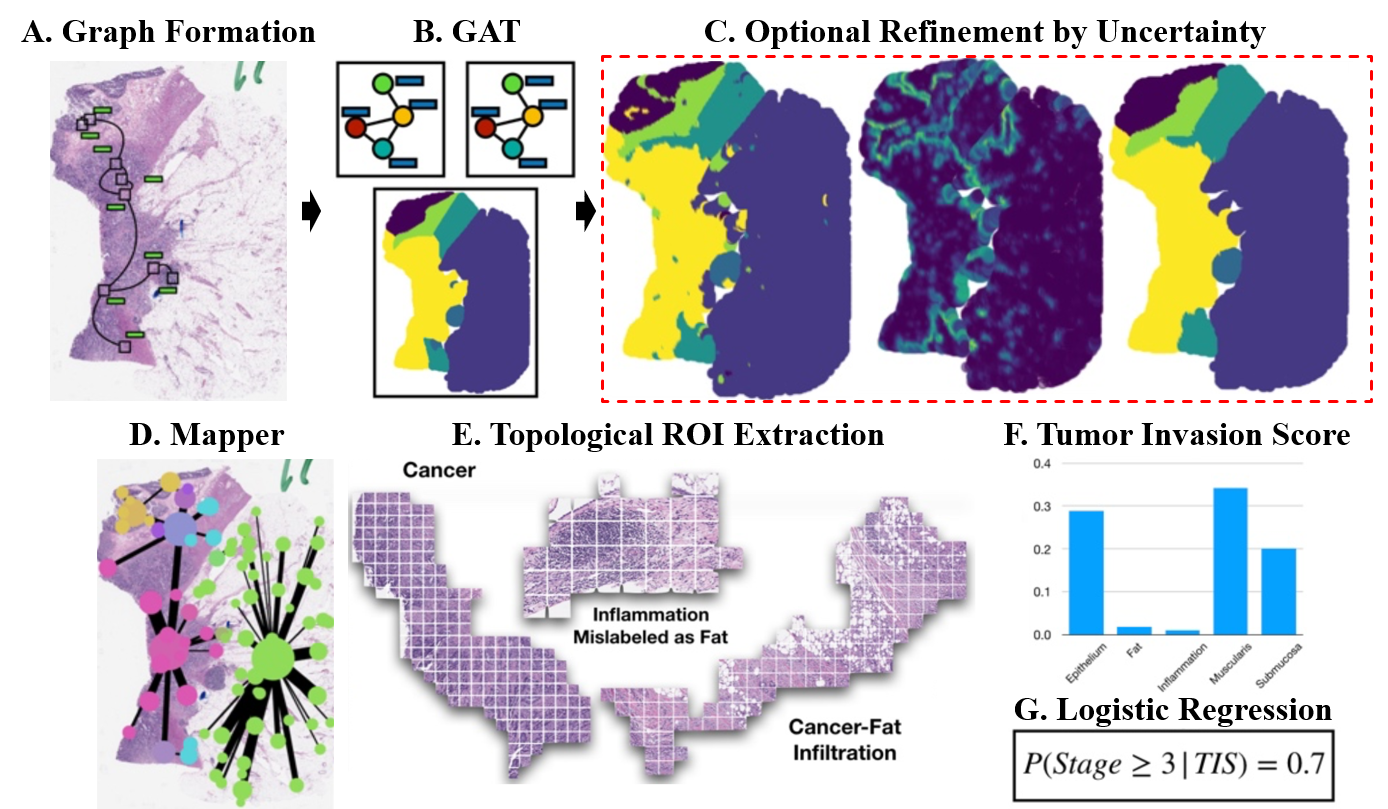}
\vspace{-10pt}
\caption{
A-C. Patch-level embeddings, graph representation and classification via a GCN. A refinement phase is incorporated through estimation of uncertainty.
D-E. The Graph Mapper summarizes high-order relationships over a WSI as a graph, where meaningful histology regions are captured.
F-G. Tumor invasion scores are used in the prediction model to form an interpretable staging score.
Image adapted from~\cite{levy2020topological}.
}
\label{fig:Fig16}
\vspace{-8pt}
\end{figure}

\begin{figure*}[!t]
\centering
\includegraphics[width=0.72\linewidth]{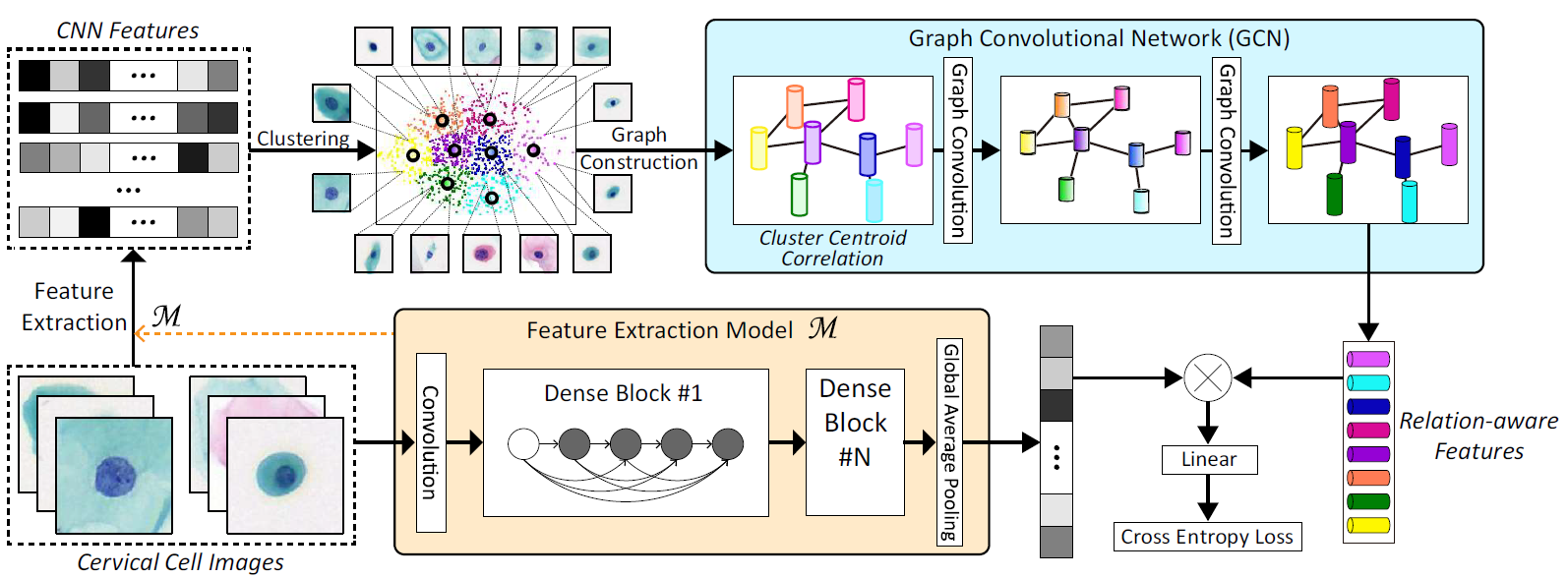}
\caption{
Classification framework for cervical cell images. Features are extracted with a CNN pre-trained on a cervical cell classification task. K-means clustering is performed on these CNN features. A graph of cluster centroid correlations is built based on intrinsic similarities, and is the input to a GCN model. The encoded representations are incorporated into the CNN features for classification. Image reproduced from~\cite{Shi2019GraphCN,shi2020cervical}.
}
\label{fig:Fig17}
\vspace{-6pt}
\end{figure*}

\subsubsection{Colorectal cancer}

Tumor staging includes both tissue and nodal stages, with higher numbers indicating a greater depth of invasion and a greater number of lymph nodes implicated in the tumor, respectively.
Levy et al.~\cite{levy2020topological} introduced a framework that used varied levels of structure to learn both local and global patterns from histological images for determining the degree of tumor invasion. Fig.~\ref{fig:Fig16} illustrates the proposed framework where the authors combined GCNs to explain the mechanisms by which tissue regions interact, and topological feature extraction methods~\cite{chazal2017introduction} to extract essential contextual information.
Patch-level classification of colon sub-compartments was conducted via a GCN as well as a refinement of patch-level predictions, in which nodes with high uncertainty were deleted, and the remaining class labels were propagated to unlabeled patches.
A topological data analysis (TDA) tool for graphs known as Graph Mapper~\cite{bodnar2020deep} was adopted as a post-hoc model explanation technique to elucidate the high-level topology of the WSI. 
The mapper generates a graph in which each node represents a cluster of WSI patches and each edge represents the degree of shared patches between the clusters. This tool can offer higher level information flow descriptors in a GNN model, substantially simplifying analysis. 
With the regions of interest (collection of patches) extracted with the mapper, the authors compute tumor invasion scores that measure the degree of overlap between the tumor and adjacent tissue region.
Finally, cancer staging is predicted via derived invasion scores using a private colon and lymph node dataset collected from the Dartmouth Hitchcock Medical Center, where the results demonstrated the potential of topological methods in the analysis of GNN models.

\begin{figure*}[!t]
\centering
\includegraphics[width=0.72\linewidth]{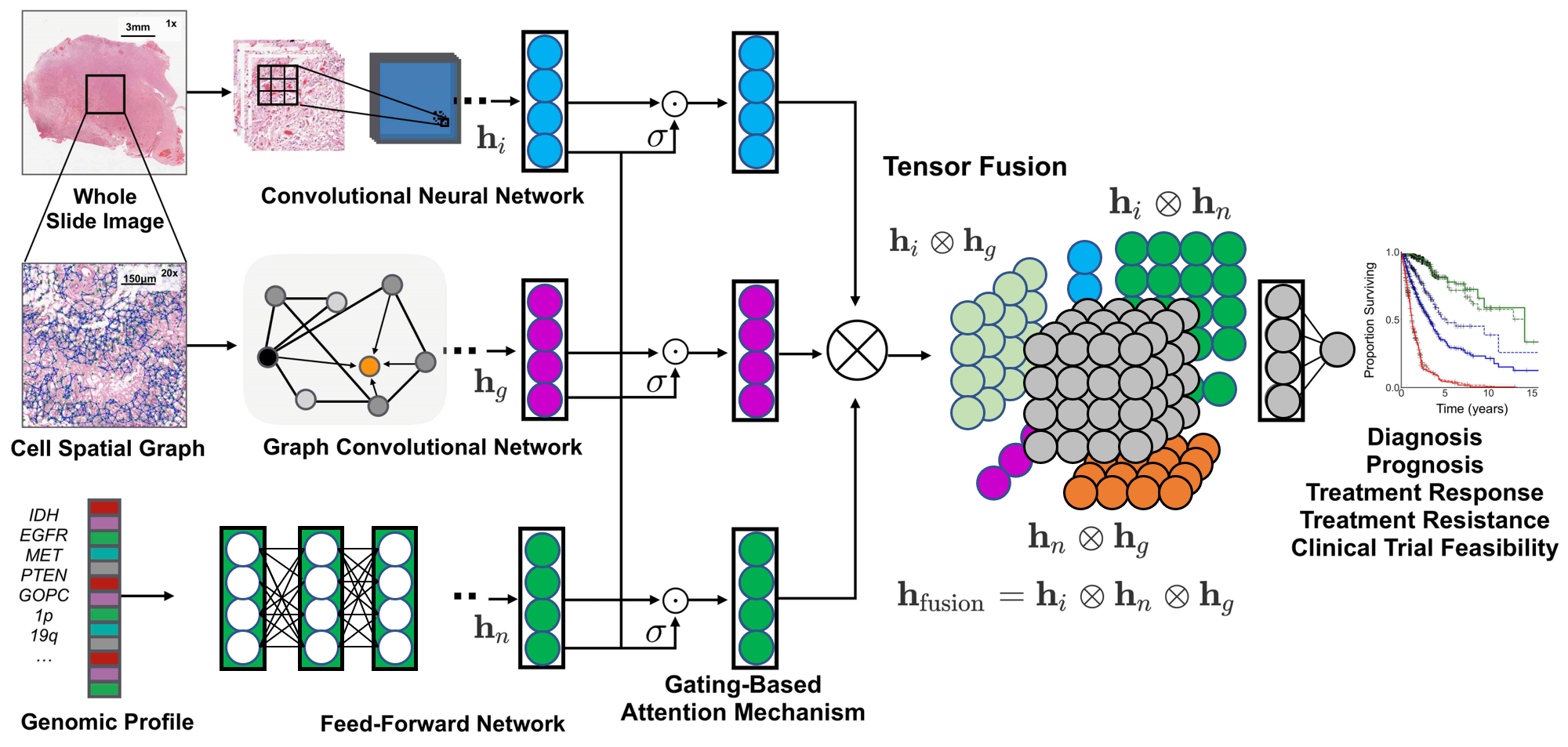}
\caption{
An integrated framework for multi-modal fusion of histology and genomics features for survival outcome prediction. Image-based features using CNNs, and graph-based features using GCNs. Recreated from~\cite{chen2020pathomic}.
}
\label{fig:Fig18}
\vspace{-6pt}
\end{figure*}

\vspace{-6pt}
\subsection{Unimodal and multi-modal feature level fusion}

In this subsection we introduce works that have used fusion techniques to extract and combined multiple rich visual representations of the same input data (unimodal fusion), or integrate information from various input modalities (multi-modal fusion) to enable more accurate and robust decisions.
The former involves integrating several feature sets acquired from different networks into a single vector, which is then used for classification. This fusion occurs in two stages: normalization of a feature, and selection of a feature.
The latter seeks to correlate and combine disparate heterogeneous modalities, such that the model can learn pairwise feature interactions and control the expressiveness of each modality. The main challenges in multi-modal data fusion are the dissimilarity of the data types being fused, and the interpretation of the results.

\subsubsection{Unimodal fusion (Cervical cancer)}
Cervical cancer is one of the most common causes of cancer death in women, and screening for abnormal cells from a cervical cytology slide is a common procedure for early detection of cervical cancer. 
In contrast with conventional CNNs which learn multi-level features through hierarchical deep architectures, Shi et al.{~\cite{Shi2019GraphCN}} combined a GCN output with deep CNN features to classify images of isolated cervical cells into five and seven classes using the SIPakMeD~\cite{plissiti2018sipakmed} and Motic (liquid-based cytology image)~\cite{shi2020cervical} datasets, respectively. 

First a CNN model pretrained for a cervical cell classification task is used to extract features of each individual cervical cell image. Then, K-means clustering is computed on the extracted features from all images to construct a graph where the centre of each cluster represents a node. The constructed graph of intrinsic similarities can be used to further investigate the potential relationships between images.
Consequently, a stacked two-layer GCN generates a relation-aware representation which is encoded into CNN features for classification, as illustrated in Fig.~\ref{fig:Fig17}. 
The authors demonstrated that the relation-aware representation generated by the GCN greatly enhances the classification performance.
Extensive experiments to validate the performance of cervical cytology classification with a GCN were also published by the same authors in~\cite{shi2020cervical}.

\subsubsection{Multi-modal fusion (Renal cancer)}

To predict clinical outcomes, oncologists often use both quantitative and qualitative information from genomics and histology{~\cite{gallego2015nonsurgical}}. However, current automated histology methods do not take genomic details into account. The following work exploits the complementary knowledge within morphological information and molecular information from genomics to better quantify tumors using graph-based methods.

Renal cell carcinoma is the most common malignant tumor of the kidney, and it is a diverse category of tumor with varying histology, clinical outcomes, and therapeutic responses. 
Renal cell carcinoma subtypes can be automatically classified through Deep learning frameworks. These algorithms can also identify features that predict survival outcomes from digital histopathological images. Several authors have used GCNs for cancer histology classification, however, its application to survival outcome prediction is less explored. 
Chen et al.~\cite{chen2020pathomic} proposed a framework for multi-modal fusion of histology and genomic features for renal cancer survival outcome prediction on the TCGA datasets (glioma and clear cell renal cell carcinoma)~\cite{tomczak2015cancer}, which contains paired whole slide images, genotype, and transcriptome data.
Their model fuses the histology image (patch features), cell-graph and genomic features into a multi-modal tensor that models interactions between the different modalities and outperforms deep learning-based feature fusion for survival outcome prediction.
This framework is illustrated in Fig.~\ref{fig:Fig18}.

The authors first extract morphological features from image-based features using CNNs, and graph-based features using GCNs, to learn cell-to-cell interactions in WSI. Cells are represented as nodes in a graph, with cells segregated using a nuclei segmentation method and connections established using KNN. 
CPC is also adopted as a self-supervised method for cell feature extraction.
The authors adopted the aggregating functions of the GraphSAGE architecture. The hierarchical self-attention pooling strategy, SAGPool~\cite{lee2019self}, is adopted to encode the hierarchical structure of cell graphs.
Then, to monitor the expressiveness of each modality, a gating-based attention system is used to perform uni-modal function fusion.
Multi-modal interpretability was considered by adopting an integrated gradient method for visualizing image saliency feature importance.

\section{Discussion and open challenges}
\label{sec:sec4}%

Beyond generating predictions relating to biology and medicine at molecular, genomic and therapeutic levels~\cite{li2021representation}, graph representation learning has also been used to support medical diagnosis through the representation of patient records as graphs by using information including brain electrical activity, functional connectivity and anatomical structures~\cite{ahmedt2021survey1}. 
As demonstrated throughout this review, graph-based deep learning has been successfully used to capture phenotypical and topological distributions in histopathology to better enable precision medicine. Numerous entity-graph based tissue representations and GNN models have been proposed for computer-aided detection and diagnosis of breast, colorectal, prostate, lung, lymphoma, skin, colon, cervical and renal cancers.

Given the utility of graphs across biomedical domains, especially to model the histology of cancer tissue, there has been a major push to exploit recent developments in deep learning for graphs in this domain. However, these applications are still in their nascent stages compared to existing research concerning conventional deep learning methods. There are challenges associated with the adoption of GNNs, and there are graph approaches yet to be explored in this domain that potentially allow a more robust and comprehensive investigation of complex biological processes that merit further investigation.
In this section, we discuss several future research directions that need to be addressed to unlock the full power of graph deep learning in digital pathology: 
1) Entity graph construction;
2) Embedding expert knowledge and clinical adoption of graph  analytics;
3) Complexity of graph models;
4) Training paradigms; and
5) Explainability of graph models.

\vspace{-6pt}
\subsection{Entity-graph construction} 
\label{subsec:sec4a}%

Defining an appropriate graph representation where vertices correspond to entities, and edges represent the connectivity of these entities, is highly relevant. 
Given a pathology task, different choices of entities in histology images can be selected as the relevant biological structures.
Several graph representations have been customized according to the relevant entity such as nuclei, tissue regions, glands or just traditional patches. However, in the majority of methods discussed in this survey, graph structures are designed manually.

\subsubsection{Pros and cons of current preprocessing steps for entity-graph construction}

\paragraph{Entity definition}
Cell-graphs have been one of the most popular graph representations, where cells are the entities used to encode cell microenvironments, including morphology of cells and cellular interactions. Such cell-graph representations were proposed in~\cite{jaume2020quantifying, jaume2020towards, sureka2020visualization, anand2020histographs, studer2021classification, zhou2019cgc, wang2020weakly, chen2020pathomic}. However, modeling a WSI as a cell-graph is non-trivial due to the large number of cells and the many possibly isolated cells and weak nuclear boundaries. This representation relies heavily on cell detection or segmentation methods. 
Although some works have used representative node sampling{~\cite{zhou2019cgc}} or agglomerative clustering{~\cite{lu2020capturing}} to remove redundancy in the graph and reduce computation cost, the majority of cell-graph based proposals assume that cell-cell interactions are the most salient sources of information. Cell-graphs do not exploit tissue macro-architectural structures, or the hierarchical nature of the tissue.

Another traditional technique for analysing WSI that include context information of ROIs is patch-graphs. Although patch-graph representations have been adopted in a number of studies~\cite{ozen2021self, aygunecs2020graph, ye2019improving, zhao2020predicting, ding2020feature, adnan2020representation, zheng2019encoding, li2018graph, wu2019weakly}, not all entities are biologically-defined and methods are limited by the patch definition. The resolution and optimal size of each image patch and the level of context offered are trade-off against one another, and are determined by the data. For example, variations in glandular morphology and size make determining an acceptable image patch size problematic. Operating at lower magnification levels may not capture cell-level features, and higher resolutions limits the ability to capture the tissue micro-environment. Thus, an automated technique that defines these patch regions and an appropriate scaling parameter from the input data is vital. 

To improve the tissue structure-function mapping, graph representations based on tissue regions have been proposed, which can also deal with one of the limitations of cell-graph as important regions may not need to only contain cells~\cite{lu2020capturing, raju2020graph, anklin2021learning}. Tissue-graphs represent well-defined tissue regions and are used to propagate information across neighboring nodes in a progressive manner at a gland or region level. Although superpixel-based approaches are proposed to address patch-graph limitations, a tissue-graph alone cannot capture local cellular information.
A combination of cell-level and patch-level features was proposed to capture local and global patterns from histological images{~\cite{chen2020pathomic}}. However, this fusion approach cannot take advantage of the hierarchy between levels.

Hierarchical graph representations were proposed as an adequate tissue representation as histological structures cannot be fully represented by cellular or tissue interactions alone. It has been shown that cell-graphs and tissue-graphs provide valuable complementary information (cellular and tissue interactions) to learn the intrinsic characteristics of cancerous tissues.
Such hierarchical analysis that captures multivariate tissue information at multiple levels has been addressed only by~\cite{zhang2020ms,pati2020hact,pati2021hierarchical,levy2020topological}. 
Nevertheless, this approach is still dependent on the construction of a cell-centered graph, which itself is limited by cell detection accuracy and is subjected to the complexity constraints of the model driven by the number of nodes.
Other works have dealt with cell detection limitations by exploiting graph wavelets with different scaling parameters{~\cite{zhang2020ms}} to obtain multilevel tissue structural information in a tissue-graph. Further, in{~\cite{levy2020topological}} micro- and macro architectures of histology images were captured with the combination of a topological data analysis tool (cell-level) and GCN (tissue-level).

\paragraph{Feature extraction}
Handcrafted and CNN-based features have been the typical methods to characterize entities. Such deep feature extraction allows use of features from a pre-trained deep architecture. However, the performance of these methods is compromised because the authors usually utilize a pre-trained model (\textit{e.g.}, trained on ImageNet) due to a lack of patch labels to fine-tune the network, and thus suffer from the domain gap between natural scene images and histopathological images. To address this limitation, a small number of works trained a feature extractor using self-supervised approaches such as CPC, VAE-GAN and auto encoder in~\cite{wang2020weakly,chen2020pathomic,zhao2020predicting,raju2020graph}.

\paragraph{Graph topology}
On current entity-graphs, each node is only connected to its spatially nearest neighbors, resulting in relatively limited information exchange during the message passing phase. Only one approach to date has computed the connections between nodes by using an adjacency learning layer in an end-to-end manner that considered the global context of all patches~\cite{adnan2020representation}.
Edge embeddings in cell-graph and tissue-graph topologies are a poorly studied field with few approaches. Learning takes place primarily at the vertices, with edge attributes serving as auxiliary information. The EGNN has only been applied in{~\cite{studer2021classification}} for colorectal cancer classification, and shows similar performance to the best model based on a 1-dimensional GNN{~\cite{morris2019weisfeiler}}.
Edge attributes can also directly inform the message passing phase operating over the vertices. In the MEGNet{~\cite{chen2019graph2}} model, vertices are updated by an aggregation of features from adjacent edges.

\subsubsection{Automated graph generation} 
Automated graph structure estimation aims to find a suitable graph to represent the data as input to the GNN model. 
By modeling graph generation as a sequential process, the graph representation (nodes, edges and embeddings) can be inferred directly from data which would be especially useful when representing tissues with a variety of complex micro- and macro environments. 
However, the majority of methods surveyed follow a standard sequential workflow which is highly dependent on the individual performance of each preprocessing step, including tissue mask detection, nuclei detection, super-pixel detection, deep feature extraction, and graph building. 
The use of neural networks to build generative graph models is gaining popularity to capture both their topology and their attributes, which can in turn lead to more robust algorithms and help to provide more accurate results. However, the effectiveness of such algorithms have not been investigated for histopathology images. Therefore, several requirements are still needed to enable the generation process.

Several works that have adopted GCNs for brain electrical activity analysis tasks {~\cite{ahmedt2021survey1}} have demonstrated that learning the graph structure from data improves classification performance in comparison to approaches where a pre-defined graph topology is used.
In digital pathology these predefined parameters per histology task are represented by fixed threshold to differentiate non-tissue pixels; patch size and number of patches for nuclei detection, and nuclei and tissue feature extraction; sample ratio of representative nuclei; thresholded KNN and distance that define topology and edges; the number of superpixels and downsampling factor per image; and selection of handcrafted features and CNN layer from which deep features are extracted. Such definitions limit the generalization of entity-graphs to different tissues, organs, and histology tasks.
Some graph generation approaches that are worthy of exploration within histopathology diagnosis are GraphGAN~\cite{wang2018graphgan}, DGMG~\cite{li2018learning}, and GCPN~\cite{you2018graph}. For instance, DGMG{~\cite{li2018learning}} can be used to generate one node at a time from each histopathology patch and then create edges one by one, to connect each node to the existing partial graph using probabilistic dependencies among nodes.

In summary, the preceding discussion exemplified the difficulties in estimating a graph structure with the desired properties from data. While there is emerging work in this field, it is ripe for further investigation. In digital pathology, automated graph generation, in which a graph model infers structural content from data, and the integration of domain knowledge, are also underutilised.

\vspace{-6pt}
\subsection{Embedding expert knowledge and clinical adoption of graph analytics} 

Incorporating domain knowledge into the model has emerged as a promising method for improving medical image analysis~\cite{xie2020survey}. 
The use of graph-based mappings with label representations (word embeddings) have been investigated to guide information propagation among nodes~\cite{chen2019multi}. For example, in basal cell carcinoma classification{~\cite{wu2019weakly}}, the embedding knowledge is represented by encoding patches based on prior expert knowledge, which bridges the gap between patch-level and image-level predictions and results in better performance.
Further, pathologists' feedback can help to improve the graph representation in terms of how to best mirror the biological relationship between cells and tissues. Thus, graph-based analysis motivates exploring the inclusion of task-specific pathological prior knowledge in the construction of the graph representations~\cite{pati2020hact}.

Another open research question is how to incorporate interdisciplinary knowledge in a principled way, rather than on a case-by-case basis.
Integrating electronic health records for personalized medicine can also boost the diagnostic power of digital pathology. The hierarchical information inherent in medical ontologies naturally lends itself to creating a rich network of medical knowledge, and other data types such as symptoms and genomics~\cite{choi2017gram}. Thus, by integrating patient records into the graph representation learning environment, tailored predictions can be generated for individual patients.

Among the AI-techniques, graph-based tissue image analysis demonstrated performance superior or comparable to domain experts in breast cancer analysis{~\cite{pati2021hierarchical}}.
These results combined with studies examining the effect of explanations on clinical end-user decisions{~\cite{jaume2020quantifying}} show generally positive results in the translation of this technology into diagnostic pathology.
Such translation will require to considered integration of standardised technologies into digital pathology workflows, resulting in an integrated approach to diagnosis and offering pathologists new tools that accelerate their workflow, increase diagnostic consistency, and reduce errors.

While, there is considerable promise for graph analytics in digital pathology, there are some challenges ahead.
These include, for example, the ability to generalize a diagnosis technique to a large population of patients which contain outliers; and to develop problem-solving skills that demand complex interactions with other medical disciplines.
Thus, more work should be conducted to investigate how a pathologist could refine a graph model decision via a human-in-the-loop system~\cite{singh2020explainable,bulten2021artificial}
Such approaches provide an important safety mechanism for detecting and correcting algorithmic errors that may occur.
A remaining challenge here is to provide frameworks with the above functionalities with reduced complexity to lower the barriers  between the systems and clinicians, to help facilitate system uptake.

Entity-graph analysis has the ability to transform pathology by providing applications that speed up workflow, improve diagnosis, and improve patient clinical outcomes. However, there is still a gap between research studies and the effort required to deliver reliable graph analytics that incorporate expert knowledge into the system, and can be integrated into existing clinical workflows.

\vspace{-6pt}
\subsection{Complexity of graph models}

Graph-based approaches for histology analysis have a high representational power, and can describe topological and geometric properties of multiple types of cancers. When compared to pixel-based approaches, the graph representation can more seamlessly describe a large tissue region. However, classical graph-based models have a high computational complexity. As a result, in the suggested learning approach, the choice of GNN architecture should be handled as a hyper-parameter.

The most common GNNs used by methods in this survey include ChebNet~\cite{defferrard2016convolutional}, GCN~\cite{kipf2017semi}, GraphSAGE~\cite{hamilton2017inductive}, GAT~\cite{velivckovic2017graph}, GIN~\cite{xu2018powerful}, and variants such as Adaptive GraphSAGE~\cite{zhou2019cgc}, RSF~\cite{such2017robust}, MS-GWNN~\cite{zhang2020ms} and FENet~\cite{ding2020feature}.
Spatial-GCNs such as GraphSAGE and GIN demonstrated their learning ability using max-, mean-, or sum-pooling aggregators. GIN has been particularly effective in computational pathology with a provably strong expressive power to learn fixed-size discriminative graph embeddings from cellular and tissue architectures in WSIs, which demonstrate translation and rotation invariance. 
However, it is noted that these GNN models inherit considerable complexity from their deep learning lineage, which can be burdensome when scaling and deploying GNNs. This is likely one of the reasons that has seen patch-based approaches remain a popular approach for many problems.

The training of GNNs remains one of the most difficult tasks due to their high memory consumption and inference latency compared to patch-based deep learning approaches.
GNNs usually require the whole graph and the intermediate states of all nodes to be saved in memory. However, the adoption of an efficient training approach is uncommon in the applications surveyed.
Various graph sampling approaches have been proposed as a way to alleviate the cost of training GNNs. Rather than training over the full graph, each iteration is run over a sampled sub-graph, whether they are sampled node-wise (GraphSage~\cite{hamilton2017inductive}), layer-wise (FastGCN~\cite{chen2018fastgcn}, $L^2$-GCN~\cite{you2020l2}), or by clustering (Cluster-GCN~\cite{chiang2019cluster}).

Some works have proposed more efficient and simple architectures that deserve attention for their potential to be adopted in computational histopathology.
The simple graph convolution (SGC)~\cite{wu2019simplifying} reduces the complexity of GCNs by repeatedly removing the non-linearities between GCN layers and collapsing multiple weight matrices into a single linear transformation. This model was adopted for emotion recognition and increased the performance speed with a comparable classification accuracy in comparison to other networks~\cite{zhong2020eeg}.
The simple scalable inception GNN (SIGN)~\cite{rossi2020sign} is explicitly designed as a shallow architecture that combines graph convolutional filters of different sizes that allow efficient pre-computation.
The efficient graph convolution (EGC)~\cite{tailor2021adaptive} method does not require trading accuracy for runtime memory or latency reductions based on an adaptive filtering approach.
GNNs can also deliver high performance for feature matching across images~\cite{sarlin2020superglue}, which can be incorporated for content-based histopathological image retrieval.

It is also important to highlight that some works exploit the cell-graph representation without the complexity of GCN processing. 
The tissue classification problem was proposed in~\cite{javed2020cellular} as a cellular community detection based on cell detection and classification into distinct cellular components (cell-graphs), and clustering of image patches (patch-level graphs) into biologically meaningful communities (specific tissue phenotype). 
The concept of constructing a graph and then using geodesic distance for community detection has outperformed deep neural networks and graph-based deep leaning methods such as ChebNet, GCNs and deep graph infomax learning (DGI)~\cite{velickovic2019deep}.

In the coming years, a key research topic will be how to effectively learn and compute GNNs in order to realise their full potential. Deep learning on graphs is inherently difficult due to the graphs' complex topological structure, which can be made up of many different types of entities and interactions. 
As such, the appropriate selection of key parameters of a model prior to representation learning is essential to capture the structural information of the histopathology slides.

\vspace{-6pt}
\subsection{Training paradigms}

As stated in previous sections, training paradigms can be divided into two main categories: training a network to learn the node embeddings used in the graph representation; and the training of the GNN model.

\subsubsection{Node embeddings}
The node embeddings are the features that are learned to represent the defined node (\textit{e.g.} cells, nucleus, patches, super-pixels). Some of the embedding features extracted through attribute networks require labeled datasets and need to be trained in a supervised manner as explained in~\cite{jaume2020quantifying,jaume2020towards,sureka2020visualization,anand2020histographs,studer2021classification,zhou2019cgc,ozen2021self,lu2020capturing,aygunecs2020graph,ye2019improving,ding2020feature,adnan2020representation,zheng2019encoding,wu2019weakly,pati2021hierarchical,pati2020hact,zhang2020ms,levy2020topological,shi2020cervical,Shi2019GraphCN}.
However, one of the main challenges in deep learning is the lack of large corpora  of manually labeled data for training, which often imposes a limitation on problems in the medical domain.
Thus, self-supervised methods are gaining interest to improve the quality of learned node embeddings~\cite{wang2020weakly,zhao2020predicting,raju2020graph,li2018graph,chen2020pathomic} by learning embedding features directly from histopathology images, rather than relying on extracting features using transfer learning, which is discussed in Subsection~\ref{subsec:sec4a}.

\subsubsection{Node/graph classification}
Training a GCN for node or graph level classification can be performed in supervised, semi-supervised or even in a self-supervised manner. 
If sufficient labels are available for nodes or graph data, the common practice is a supervised training approach, such as the methods of~\cite{jaume2020quantifying,jaume2020towards,sureka2020visualization,anand2020histographs,studer2021classification,zhou2019cgc,lu2020capturing,ye2019improving,ding2020feature,li2018graph,pati2021hierarchical,pati2020hact,zhang2020ms,levy2020topological,Shi2019GraphCN,adnan2020representation}. 

Though supervised methods can achieve high performance, they can place limitations on model complexity and can suffer when annotations are inconsistent or imprecise. 
In the absence of sufficient labeled data, weakly-supervised or semi-supervised frameworks are proposed to better capture the structure of histopathology data and reduce the human annotation workload. 
Although the issue of missing labels is not specific to the graph domain, only a few works have adopted such frameworks (pixel or patch level labels).

In semi- or weakly-supervised approach, the node embeddings are learnt from few labeled samples per class~\cite{shi2020cervical,wang2020weakly,aygunecs2020graph,wu2019weakly}.
For example, in a weakly supervised learning approach, the contributions of the individual patches to the ROI-level diagnosis are not known during training{~\cite{aygunecs2020graph}}.

In addition to the above, extensive research over past years in deep learning~\cite{kather2019deep,campanella2019clinical,hou2016patch} showed that a decision classifier based on Multiple Instance Learning (MIL) can boost the performance in classifying cancer by aggregating instance-level predictions. 
MIL only requires labels for the bag of instances rather than individual instances, which makes it well-suited for histology slide classification. One example is CLAM (Clustering-constrained attention multiple instance learning{~\cite{lu2021data}}). Even though these approaches have practical merits and can consider the important patches for predicting the staging, they do not consider the spatial relationships between patches.
Current multiple instance learning approaches using deep graphs~\cite{raju2020graph,zhao2020predicting,adnan2020representation} follow this line of research.
They can seamlessly scale to arbitrary tissue dimensions by incorporating an arbitrary number of entities and interactions, thus offering an alternative to traditional MIL{~\cite{lu2021data}}.
MIL methods can be incorporated with a GCN to take advantage of the structural information among instances{~\cite{srinidhi2020deep}}. For example, the SegGini model{~\cite{anklin2021learning}} outperforms several traditional state-of-the-art methods such as CLAM{~\cite{lu2021data}} and Context-Aware CNN (CACNN){~\cite{shaban2020context}} for weakly-supervised classification of prostate cancer.

Self supervised methods have also been successfully deployed as a training paradigm for GCNs. For example, Ozen et al.~\cite{ozen2021self} adopted a SimCLR framework~\cite{chen2020simple} along with contrastive loss to learn a representation of ROIs and perform classification.
Although the aforementioned training paradigms demonstrate remarkable performance, few works~\cite{adnan2020representation} have considered end-to-end training and the challenge that brings such as dealing with complexity of constructing a graph or labeled data, and thus this requires investigation in future works.  

Training paradigms are dependent on the availability of manually labeled data. In medical imaging and specifically histopathology obtaining a large set of labeled data is a tedious process and so weakly- and self-supervised algorithms are receiving increasing interest for learning node embeddings and performing graph classification. It is expected that in future, further research carry out on a large-scale to analyse histopathology data using GCNs in a weakly- or self-supervised manner.

\vspace{-8pt}
\subsection{Explainability of graph models}

To effectively translate graph models into clinical practise, clinicians' trust must be established. Explainability, or a model's ability to justify its outcomes and therefore assist clinicians in understanding a model's prediction, has long been seen as crucial to building trust. Understanding model behaviour beyond traditional performance indicators has thus become an important part of machine learning research, particularly in healthcare{~\cite{tonekaboni2019clinicians}}.

Explainability in deep models has focused on providing input-dependent explanations and understanding model behavior from different perspectives, including visual explanations and highlighting salient regions. We can examine the sensitivity between the input features and the predictions, for example, by looking at the gradients or weights. We can also highlight important features or regions of an image by incorporating attention mechanisms{~\cite{du2019techniques}}.
Nevertheless, compared with traditional image domains, explainability and visualization of deep learning for graphs is less explored{~\cite{ying2019gnnexplainer}}, yet explanability is critical to highlight informative structural compositions of tissue and inter-nuclear relationships, as is desired for computational histopathology.

While interpretability approaches are generally lacking within most graph network methods, it is worth noting that a few methods exist and incorporate such explanations in digital pathology as illustrated in Table~\ref{table:pathology}:
i) In~\cite{wu2019weakly} a GCN propagated supervisory information over patches to learn patch-aware interpretability in the form of a probability score.
ii) A robust spatial filtering with an attention-based architecture and node occlusion was used to capture the contribution of each nucleus and its neighborhood to the prediction~\cite{sureka2020visualization}.
iii) The Graph Mapper, a topological data analysis tool, was adopted to compress histological information to its essential structures, where meaningful histology regions are captured~\cite{levy2020topological}.
iv) In~\cite{chen2020pathomic}, an integrated gradient method was used to visualise image saliency feature importance.
v) A graph clustering visualization was used in~\cite{zhou2019cgc} to group cells with similar tissue structures.
vi) A post-hoc graph-pruning explainer, GCExplainer, was designed to identify decisive cells and interactions from the input graph~\cite{jaume2020towards}. 
vii) The gradient-based saliency method, GraphGrad-CAM, was adopted in~\cite{pati2021hierarchical} and~\cite{anklin2021learning} to measure importance scores and regions that contributed towards the classification of each class.

The majority of approaches that have incorporated explainers are limited to cell-graph analysis. Considering the pathologically aligned multi-level hierarchical tissue attributes{~\cite{pati2021hierarchical}}, the interpretability can reveal crucial entities such as nuclei, tissue parts and interactions which can mimic the pathologist's assessment and therefore, increase the level of trust between experts and AI frameworks.

Existing works however lack the definition of objectives to validate a model in terms of effective explainability, and only a single work has looked at the quality and utility of the proposed explanation methods for the intended audience (\textit{i.e.} clinicians). In~\cite{jaume2020quantifying}, the authors evaluated several graph explainers (GNNExplainer, GraphGrad-CAM, GraphGrad-CAM++, GraphLRP) to provide domain-understandable quantitative metrics based on pathologically measurable cellular properties, to make graph decisions understandable to pathologists. The authors found that at the concept-level, GraphGrad-CAM++ has the highest overall agreement with the pathologists, followed by GraphGrad-CAM and GNNExplainer.

Other methods not investigated in this survey that focus on instance-level interpretation of deep graph models that deserve attention in digital pathology for explainability at the node, edge, or node feature levels are: excitation BP~\cite{pope2019explainability}, PGM-explainer~\cite{vu2020pgm}, GraphMask~\cite{schlichtkrull2020interpreting}, Graphlime~\cite{huang2020graphlime}, and Relex~\cite{zhang2020relex}.
Other methods such as SubgraphX~\cite{yuan2021explainability} provide subgraph-level explanations which may be more intuitive and human-intelligible for digital pathology.

Knowing the subset of features from which the model outcome is derived is critical. This allows clinicians to compare model decisions with clinical judgement, which is especially useful when there is a discrepancy. It is also worth noting that clinicians expect variation in the importance of inputs to exist both across patients and populations~\cite{tonekaboni2019clinicians}. 
However, the explanations provided by methods discussed in this survey using gradient-based (GraphGrad-CAM) and perturbation-based methods (GNNExplainer) are limited to single instances. To verify and understand a deep model, pathologists need to check explanations for all input graphs, which is time-consuming and impractical. 
Models that interpret each instance independently, as previously stated, are insufficient to provide a global understanding of the trained model~\cite{guo2018explaining}. Thus, methods to provide GNN predictions on a group of instances collectively (\textit{i.e.} a population) and provide a global understanding of GNN predictions is less explored in the literature.

Instance-level methods explain GNNs with respect to each input graph, whereas model-level methods explain GNNs without regard for any specific input example. The latter specifically investigates what input graph patterns can lead to a specific GNN behaviour, such as maximising a target prediction. However, no research on interpreting GNNs at the model-level exists in digital pathology.
XGNN~\cite{yuan2020xgnn} provides model-level explanations by training a graph generator to build graph patterns that optimize a specific model prediction. The authors formulated graph creation as a reinforcement learning problem, with the graph generator predicting how to add an edge to a given graph and build a new graph at each step. The generator is then trained using a policy gradient based on feedback from the trained graph models. Several graph rules are also used to ensure that the explanations are both valid and human-readable. 
PGExplainer~\cite{luo2020parameterized} can also provide an explanation for each instance with a global view of the GNN model by incorporating a generative probabilistic model.
Nonetheless, it is unknown whether XGNN and PGExplainer can be used to perform node classification tasks for histopathology analysis, which is an important area for future research.

Given the trend of graph-based processing for a variety of applications in computational pathology, graph explainability and quantitative evaluation with a focus on clinician usability are critical. Interpretability is essential because it can aid, for example, in informed decision-making during cancer diagnosis and treatment planning. However, interpretability of GNNs within digital pathology has received insufficient attention to date.

\vspace{-6pt}
\section{Conclusion}

Through the use of whole-slide images (WSIs) and tissue microarrays (TMAs), digital pathology has transformed pathology diagnosis. The growing use of this data has also given rise to a new field of study known as computational pathology, which aims to develop machine learning techniques to provide more objective and reproducible results. 
Deep learning, in particular Convolutional Neural Networks (CNNs), have demonstrated efficacy in visual representation learning in digital pathology. To obtain image-level representations, mainstream CNN architectures typically aggregate feature representations over fixed-sized patches of the WSI. However, the patch-wise and pixel-based processing used by CNNs lacks the ability to capture global contextual information relating to meaningful entities such as cells, glands, and tissue types.
As demonstrated throughout this review, histopathology knowledge graphs enable the capture of more comprehensive and interpretable information relating to the underlying mechanisms of a disease.
Several works have attempted to adopt graph-based deep learning models to learn both local and global patterns. Entity-based analysis has the potential to improve the interpretability of deep learning techniques by identifying decisive nuclei, tissue regions and interactions. This can also potentially replicate holistic and context aware parts of a pathologist's assessment.

Our survey has provided a detailed overview of a new rapidly growing field of representation learning for computational histopathology. The enriched graph representation and learning in digital pathology has resulted in superior performance for diverse types of cancer analysis. Nevertheless, we highlight open research directions concerning the adoption of graph-based deep learning, including the explainability of graph representation learning, methods of graph construction, and the complexity of graph models and their limited training efficiency.

\vspace{-6pt}
\section{Conflict of interest statement}

The authors report no conflicts of interest.


%





\ifCLASSOPTIONcaptionsoff
  \newpage
\fi



%

{\small
        \bibliographystyle{IEEEtran}
        \bibliography{IEEEabrv,ref}
}


%








\end{document}